\documentclass{article}

\PassOptionsToPackage{numbers, compress}{natbib}

\usepackage[final]{neurips_2021}

\usepackage[utf8]{inputenc} 
\usepackage[T1]{fontenc}    
\usepackage{hyperref}       
\usepackage{url}            
\usepackage{booktabs}       
\usepackage{amsfonts}       
\usepackage{nicefrac}       
\usepackage{microtype}      
\usepackage{xcolor}         
\usepackage{graphicx}
\usepackage[resetlabels]{multibib}
\newcites{append}{Appendices References}
\usepackage{comment}

\title{Increasing Liquid State Machine Performance with Edge-of-Chaos Dynamics Organized by Astrocyte-modulated Plasticity}

\author{%
  Vladimir A. Ivanov\\
  Department of Computer Science\\
  Rutgers University\\
  Piscataway, NJ \\
  \texttt{vladimir.ivanov@rutgers.edu} \\
  \And
  Konstantinos P. Michmizos\\
  Department of Computer Science\\
  Rutgers University\\
  Piscataway, NJ \\
  \texttt{michmizos@cs.rutgers.edu} \\
}

\begin{document}

\maketitle
 \begin{abstract}
The liquid state machine (LSM) combines low training complexity and biological plausibility, which has made it an attractive machine learning framework for edge and neuromorphic computing paradigms. Originally proposed as a model of brain computation, the LSM tunes its internal weights without backpropagation of gradients, which results in lower performance compared to multi-layer neural networks. Recent findings in neuroscience suggest that astrocytes, a long-neglected non-neuronal brain cell, modulate synaptic plasticity and brain dynamics, tuning brain networks to the vicinity of the computationally optimal critical phase transition between order and chaos. Inspired by this disruptive understanding of how brain networks self-tune, we propose the neuron-astrocyte liquid state machine (NALSM)\footnote{Code and data available at \url{https://github.com/combra-lab/NALSM}} that addresses under-performance through self-organized near-critical dynamics. Similar to its biological counterpart, the astrocyte model integrates neuronal activity and provides global feedback to spike-timing-dependent plasticity (STDP), which self-organizes NALSM dynamics around a critical branching factor that is associated with the edge-of-chaos. We demonstrate that NALSM achieves state-of-the-art accuracy versus comparable LSM methods, without the need for data-specific hand-tuning. With a top accuracy of $97.61\%$ on MNIST, $97.51\%$ on N-MNIST, and $85.84\%$ on Fashion-MNIST, NALSM achieved comparable performance to current fully-connected multi-layer spiking neural networks trained via backpropagation. Our findings suggest that the further development of brain-inspired machine learning methods has the potential to reach the performance of deep learning, with the added benefits of supporting robust and energy-efficient neuromorphic computing on the edge. 

\end{abstract}

\section{Introduction}
With the recent rise of neuromorphic \cite{Kendall_2020,Tang_2020,lif_loihi,Tang_2019} and edge computing \cite{Cao_2020,Chen_2019}, the liquid state machine (LSM) learning framework \cite{Maass2002} has become an attractive alternative \cite{soures_2019,Ponghiran2019,wang_2016,Shiya_2020} to deep neural networks owing to its compatibility with energy-efficient neuromorphic hardware \cite{Li_2020,Ku_2018,Rossello_2016} and inherently low training complexity. Originally proposed as a biologically plausible model of learning, LSMs avoid training via backpropagation by using a sparse, recurrent, spiking neural network (liquid) with fixed synaptic connection weights to project inputs into a high dimensional space from which a single neural layer can learn the correct outputs. Yet, these advantages over deep networks come at the expense of 1) sub-par accuracy and 2) extensive data-specific hand-tuning of liquid weights. Interestingly, these two limitations have been targeted by several studies that tackle one \cite{balafrej2020pcritical,Brodeur_2012} or the other \cite{apstdp_bib,Norton_2006}, but not both. This has limited the widespread use of LSMs in real-world applications \cite{soures_2019}. In that sense, there is an unmet need for a unified, brain-inspired approach that is directly applicable to the emerging neuromorphic and edge computing technologies, facilitating them to go mainstream.

As a general heuristic, LSM accuracy is maximized when LSM dynamics are positioned at the edge-of-chaos \cite{Legenstein2007,Bertschinger2004,Langton1990} and specifically in the vicinity of a critical phase transition \cite{Shew2011,de_Arcangelis2010,Kinouchi2006,Haldeman_2005} that separates: 1) the sub-critical phase, where network activity decays, and 2) the super-critical (chaotic) phase, where network activity gets exponentially amplified. Strikingly, brain networks have also been found to operate near a critical phase transition \cite{Beggs2003,Shriki2013,Chialvo2010} that is modeled as a branching process \cite{Haldeman_2005,Beggs2003}. Current LSM tuning methods organize network dynamics at the critical branching factor by adding forward and backward communication channels on top of the liquid \cite{balafrej2020pcritical,Brodeur_2012}. This, however, results in significant increases in training complexity and violates the LSM's brain-inspired self-organization principles. For example, these methods lack local plasticity rules that are widely observed in the brain and considered a key component for both biological \cite{Feldman2009} and neuromorphic learning \cite{lif_loihi,Tang_2020,Tang_2019}. A particular local learning rule, spike-timing-dependent plasticity (STDP), is known to improve LSM accuracy \cite{apstdp_bib,Norton_2006}. Yet, current methods of incorporating STDP into LSMs further exacerbate the limitations of data-specific hand-tuning as they require additional mechanisms to compensate for the STDP-imposed saturation of synaptic weights \cite{apstdp_bib,stepp2015,Zenke2017,Watt2010,Abbott2000}. This signifies the scarcity of LSM tuning methods that are both computationally efficient and data-independent.

A long-neglected non-neuronal cell in the brain, astrocytes, is now known to play key roles in modulating brain networks \cite{RN46, Han2013, Tewari2013,RN52,RN50,Petrelli_2020}, from modifying synaptic plasticity \cite{Manninen_2020,Sibille_2014,Min_2012} to facilitating switching between cognitive states \cite{Thrane2012, Foley2017,Bojarskaite2019,Ingiosi2019} that have been linked to a narrow spectrum of dynamics around the critical phase transition \cite{Tagliazucchi2016,Bellay2015,Hahn2017,Priesemann2013,Fagerholm2015}. The mechanisms that astrocytes use to modulate neurons include the integration of the activity of thousands of synapses into a slow intracellular continuous signal that feeds back to neurons by affecting their synaptic plasticity \cite{Parpura8629,RN62,RN63,RN64,Min_2012}. The unique spatio-temporal attributes \cite{RN48,RN67} identified in astrocytes align well with the brain's remarkable ability to self-organize its massive and highly recursive networks near criticality. That is why astrocytes present a fascinating possibility of forming a unified feedback modulation mechanism required to improve baseline LSM accuracy while eliminating data-specific hand-tuning.

Here, we propose the neuron-astrocyte liquid state machine (NALSM), where a biologically inspired astrocyte model organized liquid dynamics near a critical phase transition, by modulating STDP. We show that NALSM combined the computational benefits of both STDP and critical branching dynamics by demonstrating its accuracy advantage compared to other LSM methods on two datasets: 1) MNIST \cite{mnist_bib}, and 2) N-MNIST \cite{nmnist_bib}. We demonstrate that, similar to its biological counterpart that handles new and unstructured information with robustness and versatility, NALSM maintains the state-of-the-art LSM performance without re-tuning training parameters for each tested dataset. We also show that a NALSM with a large enough liquid can attain comparable accuracy to fully-connected multi-layer spiking neural networks trained via backpropagation on 1) MNIST \cite{mnist_bib}, 2) N-MNIST \cite{nmnist_bib}, as well as 3) Fashion-MNIST \cite{Xiao2017}. Our results suggest that the under-performance and high training difficulty of current neuromorphic methods can be addressed by harvesting neuroscience knowledge and further translating biological principles to computational mechanisms.

\section{Methods} \label{methods_main}
\subsection{The neuron-astrocyte liquid state machine} \label{methods_models}

To construct the NALSM, we started with a baseline LSM model consisting of 2 layers: 1) a spiking liquid, and 2) a linear output layer. Next, we added STDP to the LSM liquid, forming the LSM+STDP model. We developed a biologically faithful leaky-integrate-and-modulate (LIM) astrocyte model, which we embedded in the LSM+STDP liquid, to form the NALSM. The process is formalized below.

\paragraph{LSM Model}
We implemented the baseline LSM as a 3-dimensional neural network (liquid) consisting of $1,000$ neurons surrounded by 1-dimensional layers of input and output neurons. Number of input neurons was $784$ and $2,312$ for MNIST and N-MNIST, respectively (See Appendix \ref{sup_datasets}). We used the leaky-integrate-and-fire (LIF) model \cite{lif_loihi} for input and liquid neurons, modeled as:  
\begin{equation}\label{eq:neuron_voltage}
    \frac{dv_{i}}{dt} = -\frac{1}{\tau_{v}}v_{i}(t) +u_{i}(t)-\theta_{i}\sigma_{i}(t)\\
\end{equation}
\begin{equation}\label{eq:synaptic_decay}
    u_{i}(t) = \sum_{j\neq i} w_{ij}(\alpha_{u}*\sigma_{j})(t)+b_{i}\\ 
\end{equation}
where $v_i$ is the membrane potential and $u_i$ is the synaptic response current of neuron $i$, $\theta_{i}$ is the membrane potential threshold, $\sigma_{i}(t)=\sum_{k}\delta(t-t_{i}^{k})$ is the spike train of neuron $i$ with $t_{i}^{k}$ being the time of the $k$-th spike, $w_{ij}$ is the weight connecting neuron $j$ to $i$, $b_{i}$ is the bias of neuron $i$, and $\alpha_{u}(t)=\tau_{u}^{-1} \exp(-t/\tau_{u})H(t)$ is the synaptic filter with H(t) being the unit step function (See Appendix \ref{sup_neuron_params}). All LIF neurons had a $2$ $ms$ absolute refractory period. Liquid neurons were excitatory and inhibitory with $80\% / 20\%$ ratio. Input neurons did not have an excitatory/inhibitory distinction and had random excitatory and inhibitory connections to liquid neurons. From here on, we will refer to connections between input neurons and liquid neurons as IL connections, inter-liquid connections as LL, and liquid to output connections as LO. In line with \cite{Maass2002}, we created LL connections using probabilities based on Euclidean distance, $D(i,j)$, between any two neurons $i$, $j$:
\begin{equation}\label{eq:connection_probability}
    P(i,j) = C \cdot exp\left(-\left(\frac{D(i,j)}{\lambda}\right)^{2}\right)\\
\end{equation}
with closer neurons having higher connection probability. Parameters $C$ and $\lambda$ set the amplitude and horizontal shift, respectively, of the probability distribution (See Appendix \ref{sup_connectivity}). Density of IL connections was 15\%. The output layer was a dense layer consisting of $10$ linear neurons.

\paragraph{LSM+STDP Model}
We added unsupervised, local learning to the LSM model by letting STDP change each LL and IL connection \cite{Morrison2008}, modeled as:
\begin{equation}\label{eq:STDP_W_CHANGE}
\frac{dw}{dt} = A_{+}T_{pre}\sum_{o}\delta (t-t^{o}_{post}) - A_{-}T_{post}\sum_{i}\delta (t-t^{i}_{pre})\\
\end{equation}
where $A_{+}=A_{-}=0.15$ are the potentiation/depression learning rates and $T_{pre}$/$T_{post}$ are the pre/post-synaptic trace variables, modeled as,
\begin{equation}\label{eq:stdp_pre_trace}
\tau_{+}^{*} \frac{d T_{pre}}{dt} =-T_{pre}+a_{+}\sum_{i}\delta (t-t_{pre}^{i})\\
\end{equation}
\begin{equation}\label{eq:stdp_post_trace}
\tau_{-}^{*} \frac{d T_{post}}{dt} =-T_{post}+a_{-}\sum_{o}\delta (t-t_{post}^{o})\\
\end{equation}
where $a_{+}=a_{-}=0.1$ are the discrete contributions of each spike to the trace variable, $\tau_{+}^{*}=\tau_{-}^{*}=10$ $ms$ are the decay time constants, $t_{pre}^{i}$ and $t_{post}^{o}$ are the times of the pre-synaptic and post-synaptic spikes, respectively. We constrained connection weights to: 1) IL: $[-3,3]$ , 2) excitatory LL: $[0,3]$, and 3) inhibitory LL: $[-3,0]$. We used the same STDP parameters for all models and experiments.

\paragraph{LIM Astrocyte Model}
We developed the astrocyte model as a leaky integrator with a continuous output value $A_{-}^{astro}$, expressed as:
\begin{equation}\label{eq:lim_astro}
\tau_{asto} \frac{dA_{-}^{astro}}{dt}=-A_{-}^{astro} +w_{astro}\sum_{i\in N_{liq}}\delta(t-t_{i})-w_{astro}\sum_{j\in N_{inp}}\delta(t-t_{j})+b_{astro}\\
\end{equation}
where $A_{-}^{astro}$ directly mapped to $A_{-}$ in equation (\ref{eq:STDP_W_CHANGE}), $b_{astro}=A_{+}$ adjusted the astrocyte output to the fixed STDP potentiation learning rate, $N_{liq}$ and $N_{inp}$ are the sets of liquid and input neurons, respectively, and $w_{astro}$ set astrocyte responsiveness to network activity (See Appendix \ref{sup_astro}). Ignoring the decay and bias terms, the astrocyte model computed the difference in the number of spikes produced by liquid neurons and input neurons. Functionally, this is equivalent to computing the ratio of spikes emitted by the liquid over the input neurons:
\begin{equation}\label{eq:bf_proxy}
BF_{proxy}(t)=\frac{\sum_{i\in N_{liq}}\delta(t-t_{i})}{\sum_{j\in N_{inp}}\delta(t-t_{j})}\\
\end{equation}

\begin{figure}
  \centering
    \includegraphics[width=1.0\linewidth]{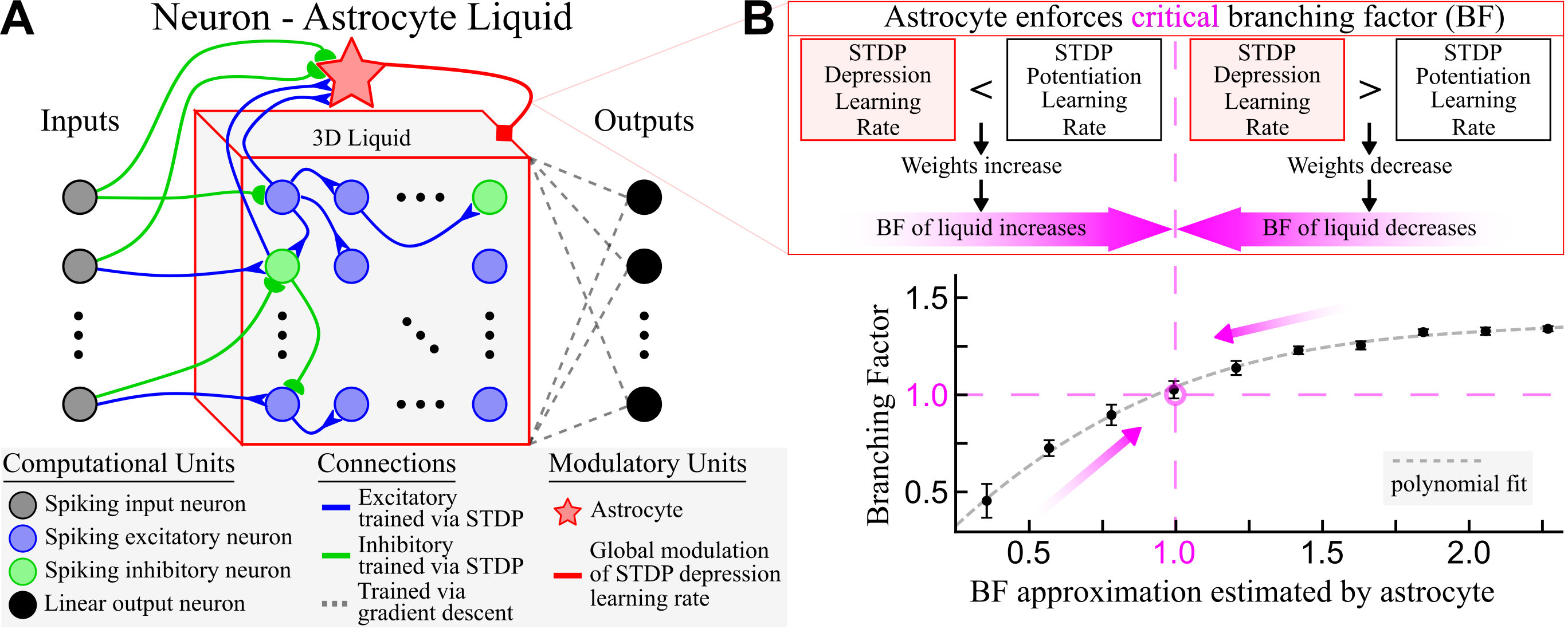}
  \caption{{\bf{NALSM architecture and astrocyte modulation of liquid dynamics.}}
  ( {\bf{A}} ) The neuron-astrocyte liquid was modeled as a 3-dimensional network of excitatory and inhibitory spiking neurons connected with sparse, recurrent connections with spike-timing-dependent plasticity (STDP). Input neurons projected excitatory and inhibitory connections to the liquid. Receiving each liquid neuron's spike count per input sample, a dense linear output layer was trained via gradient descent to classify inputs. ( {\bf{B}} ) To organize liquid dynamics at the critical branching factor, an astrocyte integrated input and liquid neuron activity and, in turn, set the global STDP depression learning rate. Data points are binned averages over `BF approximation' metric. Error bars are standard deviation. See Appendix \ref{appendix_curve_fits} for polynomial fits.
  }
  \label{methods_fig_0}
\end{figure}

Specifically, when the liquid produced more spikes than the input neurons, their difference was positive which translated to $BF_{proxy}>1.0$, and vice versa. This approach to measure liquid dynamics acted as a network level approximation of the branching factor, $\sigma_{BF}$, which is normally evaluated for each neuron (See \ref{methods_BF}). We empirically confirmed that $BF_{proxy}=1.0$ aligned with the critical branching factor, $\sigma_{BF}=1.0$ (Fig. \ref{methods_fig_0} B). Hence, as dynamics became progressively super-critical ($\sigma_{BF}>1.0$), $BF_{proxy}$ became greater than $1$, which caused the LIM astrocyte to increase STDP depression learning rate above the fixed STDP potentiation learning rate ($A_{-}^{astro}\rightarrow A_{-}>A_{+}$). This caused STDP to decrease the average weight of LL and IL connections, which decreased number of spikes produced by the liquid and made dynamics less super-critical (Fig. \ref{methods_fig_0} B). The reverse occurred as dynamics became progressively sub-critical. As a result of astrocyte modulation, liquid dynamics oscillated between sub-critical and super-critical until eventual stabilization near the critical branching factor (See Appendix \ref{sup_astro}).

\paragraph{NALSM Model}
We completed NALSM by adding the LIM astrocyte to the LSM+STDP model's liquid (Fig. \ref{methods_fig_0} A). As described above, the LIM astrocyte integrated activity from input and liquid neurons, and continuously controlled the STDP depression learning rate.

\subsection{Training} \label{training_methodology_main}

Model training was done in $3$ steps: 1) initialization of IL and LL liquid connections, 2) passing all data through the liquid resulting in liquid neuron spike counts, and 3) training the output layer on the spike counts. The steps are further detailed below.

\subsubsection{Liquid initialization} \label{methods__liquid_initialization}

\paragraph{LSM} 
We initialized all IL and LL connections with a single weight value, maintaining originally defined connection signs \cite{Zhang2019}. Weights were constant during spike count collection.

\paragraph{LSM+STDP}
We initialized IL and LL connections with STDP by consecutively presenting all the training images to the liquid. Starting with initially maximal connections, $3(-3)$ for excitatory(inhibitory) connections, we let STDP continuously adjusted weights while presenting the liquid with a randomly ordered series of MNIST training image snapshots, each lasting $20$ $ms$. For N-MNIST, we randomly sampled each $20$ $ms$ snapshot from the $0-250$ $ms$ range, as a way to account for the variability in the temporal dimension (See \ref{methods_experiments}). In each case, we used a total of $50,000$ snapshots, each corresponding to a unique training image. We used STDP only for weight initialization. Initialized weights were fixed during spike count collection. 

\paragraph{NALSM}
We used the LSM+STDP weight initialization process with the exception of added STDP modulation by the LIM astrocyte. We used this set of initialized weights as the starting point for each sample in the spike counting phase, during which astrocyte-modulated STDP continued to adjust synaptic weights to compensate for slight deviations in dynamics caused by each input sample's different level of activity. For each sample, parameters $A_{+}$ from (\ref{eq:STDP_W_CHANGE}) and $b_{astro}$ from (\ref{eq:lim_astro}) were both initialized to $0.15$ and decayed at a rate of $0.99$ for the duration of sample input.

\subsubsection{Output layer training} \label{methods__output layer}

We assembled spike counts by presenting each sample image to the liquid for $250$ $ms$ and counting the number of spikes emitted by each liquid neuron for the full duration of input. We used Adam optimizer to batch train the output layer on spike count vectors by minimizing the cross entropy loss with L2-regularization,
\begin{equation}\label{eq:207}
    \mathcal{L}(y_{i},\hat{y}) = -\frac{1}{m} \sum^{m}_{i=1} y_{i}log(\hat{y}_{i}) + (1-y_{i})log(1-\hat{y}_{i}) + \frac{\lambda_{reg}}{2m}||W_{out}||^{2}_{F}
\end{equation}
where $m=250$ is the batch size, $W_{out}$ is the output layer weight matrix, $\lambda_{reg}=5\times 10^{-10}$ is the regularization hyperparameter, $y_{i}$ and $\hat{y}_{i}$ are the normalized vectors denoting the predicted label and the target label, respectively. Prior to training, we initialized output layer weights/biases to $0.0$, and the learning rate to $0.1$. We trained the output layer until validation accuracy peaked (up to a maximum of $5,000$ epochs), at which point we evaluated model test accuracy.

\subsection{Experiments} \label{methods_experiments}
We performed all LSM comparison experiments on MNIST and N-MNIST datasets (See \ref{sup_datasets}). Using 10 randomly generated networks for each dataset, we trained 1) the baseline LSM model, 2) the LSM+STDP model, 3) the NALSM model (See \ref{methods_models}), and 4) the LSM+AP-STDP model, a method for incorporating STDP in the LSM liquid \cite{apstdp_bib} (See Appendix \ref{sup_apstdp}). First, we evaluated LSM model accuracy with respect to liquid weight, which ranged in $0.4-1.2$ for MNIST and $0.8-1.35$ for N-MNIST. We used a random seed for each training session. Next, we evaluated corresponding network dynamics of each network/weight combination by measuring the liquid's branching factor on 20 randomly sampled inputs (See \ref{methods_BF}). To have comparable results for each network, we trained the remaining models using the same seed that resulted in peak LSM accuracy. For NALSM, we used the same initialization and parameters for all networks and datasets. For LSM+AP-STDP, we hand-tuned STDP control parameters for each network and dataset combination to maximize validation accuracy (See Appendix \ref{sup_apstdp}). Additionally, we trained NALSM on Fashion-MNIST dataset (See \ref{sup_datasets}) using the same 10 randomly generated networks that we had used for MNIST.

\paragraph{Sparse neuron-astrocyte connectivity}
We tested NALSM's accuracy as a function of neuron-astrocyte connection density on $3$ best performing networks (per dataset). Keeping the proportion of neurons sampled by the astrocyte the same for both input neurons and liquid neurons, we trained NALSM with $10\%$, $20\%$, $40\%$, $60\%$, and $80\%$ neuron-astrocyte density over $3$ seeds for each of $3$ networks. Regardless of connection sparseness, all IL/LL connections were modulated by $A_{-}^{astro}$ (See \ref{methods_models})

\paragraph{NALSM with larger liquid sizes}
We tested NALSM performance for larger liquids. For each size, we trained $3$ randomly generated networks, each on a random seed. All parameters and initialization were same as for $1,000$ neuron liquid. For maximum accuracy, we trained NALSM with an $8,000$ neuron liquid. For each dataset, we used $5$ randomly generated networks trained on a random seed. Parameter $w_{astro}=0.0075$ for all datasets. All other parameters and initialization were as before.   

\subsubsection{Branching factor of liquid} \label{methods_BF}
To evaluate a liquid's dynamics, we used the network branching factor, $\sigma_{BF}$, which quantifies network information flow amplification/decay. Liquid dynamics are sub-critical, near-critical and super-critical when $\sigma_{BF} < 1.0$, $\sigma_{BF} \approx 1.0$, and $\sigma_{BF} > 1.0$. We calculated $\sigma_{BF}$ as done in \cite{stepp2015}, with offset $\phi=0$ and time window $\Delta = 4$ $ms$ as per \cite{Beggs2003}.

\subsubsection{Kernel quality of liquid} \label{sup_kernel_quality}
We evaluated liquid 1) linear separation, and 2) generalization capability using methods from \cite{Maass_2004}. For the MNIST and N-MNIST test sets, we computed the rank of matrix $M$ assembled from $k$ randomly selected spike count vectors, resulting in shape $N_{liq}\times k$. We repeated this on $1,000$ shuffles of spike vectors. For linear separation, we used spike counts from model testing phase. For generalization capability, we added noise to input data and evaluated new spike count vectors (See \ref{methods__output layer}). For MNIST, we added $\mathcal{N}(0,125)$ noise to each pixel value. For N-MNIST, we time-shifted each event by $\mathcal{N}(0,10)$. Taken together across all models and both datasets, we rescaled ranks of each measure to $0-1$ range and subtracted measure $1$ from measure $2$ as in \cite{Maass_2004}. Due to negative differences, we again rescaled all differences to $0-1$.

\begin{figure}[b!]
  \centering
    \includegraphics[width=1.0\linewidth]{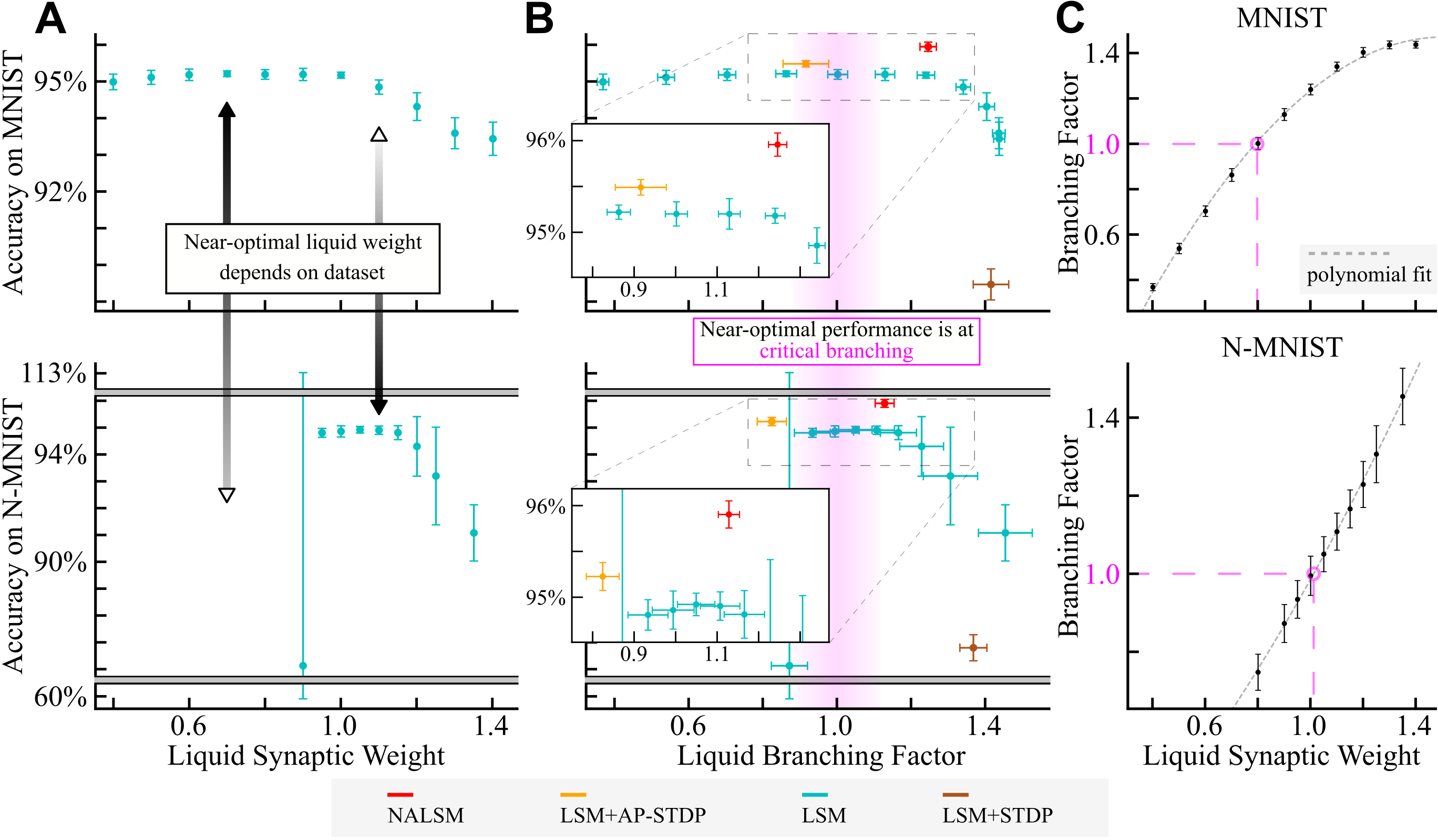}
  \caption{{\bf{LSM accuracy depended on liquid weight and dynamics.}}
  ( {\bf{A}} ) LSM accuracy shown as a function of its liquid weight, averaged over a set of $10$ randomly generated networks for MNIST and N-MNIST datasets. ( {\bf{B}} ) LSM accuracy shown with respect to liquid dynamics set by liquid synaptic weight. For each weight, liquid dynamics were measured and averaged over all $10$ networks. Similarly, accuracy and resulting liquid dynamics are shown for each model: 1) NALSM, 2) LSM+AP-STDP, and 3) LSM+STDP. ( {\bf{C}} ) Liquid dynamics shown with respect to liquid weight, averaged over all $10$ networks. Error bars are standard deviation. See Appendix \ref{appendix_curve_fits} for polynomial fits.  
  }
  \label{results_fig_1}
\end{figure}

\section{Results}
\subsection{Baseline LSM performance}
We established a benchmark accuracy for the baseline LSM on MNIST and N-MNIST datasets (See \ref{methods_experiments}). We acquired our baseline by averaging over 10 randomly generated liquids with $1,000$ neurons (See \ref{methods_models}). The LSM achieved a top accuracy of $95.44\%$ ($95.30 \pm 0.11\%$) on MNIST, and $95.35\%$ ($95.02 \pm 0.15\%$) on N-MNIST (See \ref{training_methodology_main}). For MNIST, this was comparable to the previously reported state-of-the-art LSM accuracy \cite{multi_liquid_lsm_2019}, using the same sized liquid. Further, LSM accuracy was very sensitive to the liquid's weight (Fig. \ref{results_fig_1} A).

\subsection{LSM performance peaked at the critical branching factor}
The peak LSM accuracy on each dataset corresponded to a different liquid synaptic weight. Specifically, there were cases where a liquid with weights tuned for maximum accuracy on MNIST, would catastrophically fail on N-MNIST (Fig. \ref{results_fig_1} A). Also, LSM accuracy on MNIST plateaued for a wider range of weights than on N-MNIST, which can be attributed to N-MNIST's greater difficulty caused by its variability over the temporal dimension (See Appendix \ref{sup_datasets}). Taken together, this indicated that LSM training requires extensive hand-tuning of weights for each specific dataset.

Since critical dynamics are well known to result in near-maximum LSM performance \cite{boedecker_2012,Legenstein2007,Bertschinger2004,Langton1990}, dataset-specific hand-tuning can be significantly reduced by replacing accuracy with liquid dynamics as the target output of weight tuning. Indeed, LSM accuracy was near-maximum for both datasets, when the liquid's branching factor was in $1.0-1.2$ range, or slightly super-critical (Fig. \ref{results_fig_1} B) (See \ref{methods_BF}). This agrees with studies showing that information transfer in finite sized systems peaks at slightly super-critical dynamics \cite{Ribeiro_2007}. Although each dataset still had different weight ranges corresponding to the critical branching factor, the relationship between liquid dynamics and weight was positive for both datasets (Fig. \ref{results_fig_1} C). Known to generalize beyond specific datasets \cite{Bertschinger2004}, this relationship suggested that near-critical dynamics can be organized using STDP, by providing it directional feedback from current liquid dynamics.

\subsection{Astrocyte-modulated plasticity organized liquid dynamics near criticality}
The LIM astrocyte model stabilized liquid dynamics near the critical branching factor as we presented a continuous stream of samples to the neuron-astrocyte liquid (Fig. \ref{results_fig_1} B). The NALSM's slightly super-critical stabilization suggested that liquid dynamics were at the edge-of-chaos. While chaotic activity is known to correspond to super-critical branching dynamics in some cases \cite{Haldeman_2005}, such correspondence is not guaranteed. Hence, we examined additional network properties that are necessary and indicative of chaotic activity (See Appendix \ref{appendix_edge_of_chaos}). Specifically, the astrocyte-modulated liquid had coexistence of small and large synaptic weights (Fig. \ref{sup_fig_input_liquid_weights}), as well as a balance of excitation and inhibition, both of which are necessary for the existence of chaotic network activity \cite{kus_2020,Vreeswijk_1996}. Further supporting a chaotic activity, the neuron-astrocyte liquid spike activity appeared irregular (Fig. \ref{sup_fig_input_liquid_spike_raster}). We also performed autocorrelation analysis on liquid neuron spike trains, which further suggested the existence of chaotic activity with a correspondence between edge-of-chaos dynamics and critical branching dynamics (Fig. \ref{sup_fig_input_liquid_autocorrelation}) \cite{Ostojic_2014,Rajan_2010} (See Appendix \ref{appendix_edge_of_chaos}). Given that liquid dynamics were directly approximated by the LIM astrocyte, which directly controlled the STDP depression learning rate (See \ref{methods_models}), NALSM required no dataset-specific hand-tuning. As a result, we used the same weight initialization and parameters to benchmark NALSM (See \ref{methods_experiments}).

\begin{figure}[b]
  \centering
    \includegraphics[width=1.0\linewidth]{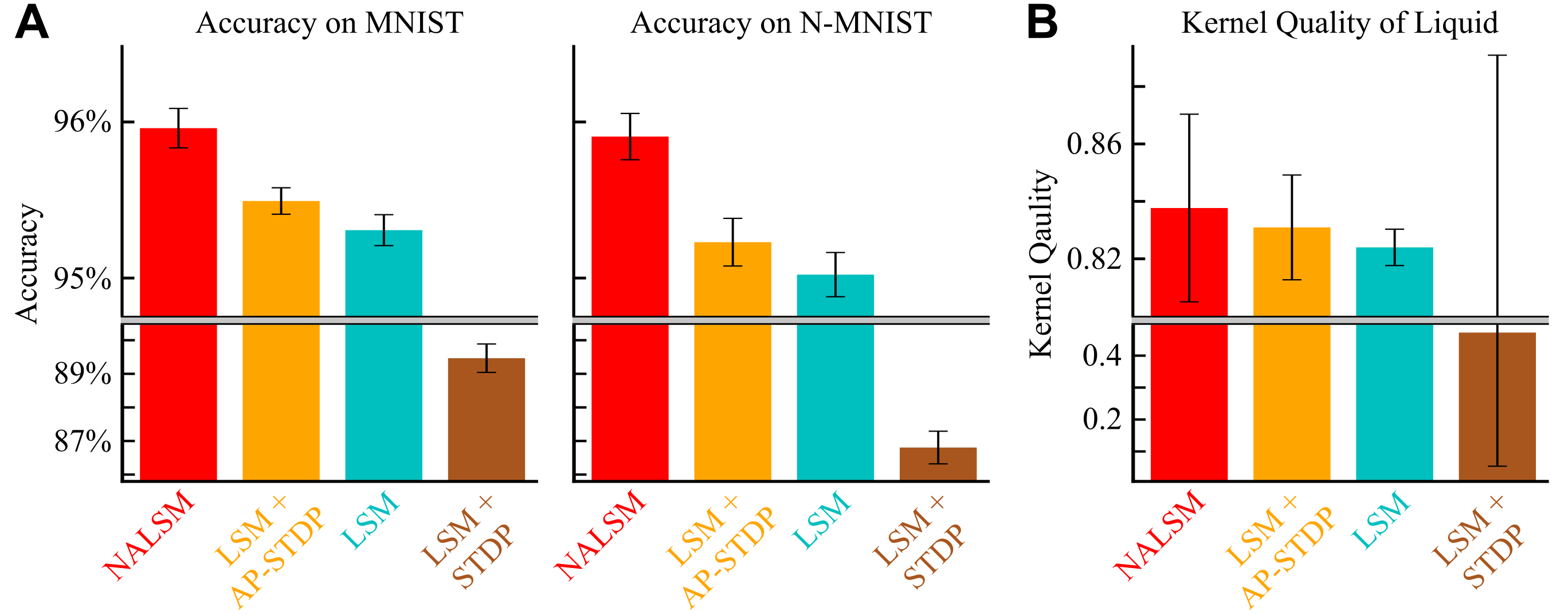}
  \caption{{\bf{Comparison of model accuracy and liquid computational capacity.}}
  ( {\bf{A}} ) Accuracy performance of the proposed NALSM model was compared, on MNIST and N-MNIST, against $3$ related models: 1) the baseline LSM, 2) LSM with activity-based STDP (LSM+AP-STDP), and 3) LSM with unregulated STDP (LSM+STDP). For each dataset and model, accuracy was evaluated using $10$ randomly generated networks, each of which was trained on a random seed. This set of $10$ seeds was used for all models. ( {\bf{B}} ) Computational capacity of each model was measured using a kernel quality metric that encompassed the linear separation and generalization capability of the liquid (See \ref{sup_kernel_quality}). Error bars are standard deviation.
  }
  \label{results_fig_2_acc_kq}
\end{figure}

\begin{figure}[b]
  \centering
    \includegraphics[width=1.0\linewidth]{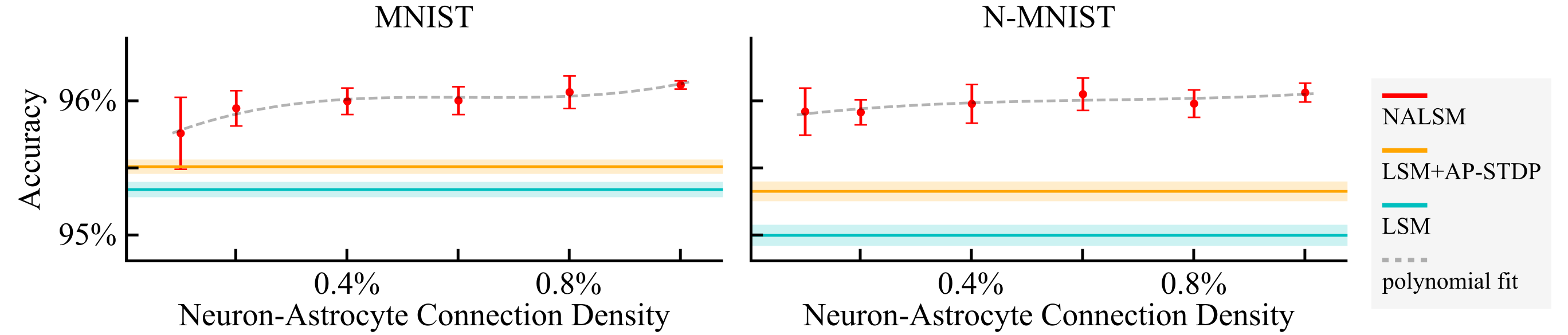}
  \caption{{\bf{NALSM maintains accuracy advantage with sparse neuron-astrocyte connectivity.}} For MNIST and N-MNIST, NALSM accuracy was evaluated with respect to neuron-astrocyte connection density. For each density, NALSM performance was compared to LSM and LSM+AP-STDP average accuracy. NALSM data points are average values over $9$ experiments ($3$ networks $\times$ $3$ seeds). Error bars and shaded areas are standard deviation. See Appendix \ref{appendix_curve_fits} for polynomial fits.}
  \label{results_fig_3_ABLATION}
\end{figure}

\subsection{Benchmarking NALSM performance on MNIST and N-MNIST}
On both datasets, NALSM achieved superior performance to comparable LSM models of the same size. Using $1,000$ liquid neurons, NALSM achieved a top accuracy of $96.15\%$ ($95.96 \pm 0.13\%$) on MNIST and $96.13\%$ ($95.90 \pm 0.16\%$) on N-MNIST; outperforming LSM model's top accuracy by $0.71\%$ on MNIST and $0.78\%$ on N-MNIST (Fig. \ref{results_fig_2_acc_kq} A). We also compared NALSM to a state-of-the-art LSM STDP method, AP-STDP (See \ref{methods_experiments}). The LSM+AP-STDP model required more extensive dataset-specific hand-tuning than the baseline LSM due to its additional STDP control parameters (See Appendix \ref{sup_apstdp}). Resulting in top accuracy of $95.62\%$ ($95.49 \pm 0.09\%$) and $95.43\%$ ($95.23 \pm 0.16\%$), the LSM+AP-STDP model was superseded by NALSM by $0.53\%$ and $0.70\%$ on MNIST and N-MNIST, respectively. As a control measure, we also trained a LSM with unregulated STDP (See \ref{methods_models}). The LSM+STDP model significantly under-performed compared to all other models achieving a top accuracy of $90.52\%$ ($89.47 \pm 0.45\%$) on MNIST and $87.71\%$ ($86.80 \pm 0.51\%$) on N-MNIST. We attributed this under-performance to the LSM+STDP liquid's excessive super-critical dynamics (Fig. \ref{results_fig_1} B) which are well known to decrease liquid computational capacity \cite{boedecker_2012,Ribeiro_2007}.   

The NALSM had the most robust accuracy performance across the two datasets out of all the compared LSM models. With no dataset-specific tuning, NALSM's average accuracy on N-MNIST was lower than the accuracy on MNIST by only $-0.05\%$. This was $5-50$ times less than for the LSM+AP-STDP ($-0.26\%$), LSM ($-0.29\%$), and LSM+STDP ($-2.66\%$) models.

We attributed the NALSM's performance advantage to the improved computational properties of its liquid. For both tested datasets, the NALSM achieved slightly super-critical branching dynamics where baseline LSM performance peaked right before it started to decline with increasing super-critical dynamics (Fig. \ref{results_fig_1} B). This suggested that NALSM's performance advantage, compared to a LSM with similar dynamics, was due to the addition of astrocyte-modulated STDP (See \ref{methods_models}, \ref{methods__liquid_initialization}). While LSM+AP-STDP and LSM+STDP models also had STDP, their lower performance can be explained by their excessively sub-critical and super-critical dynamics, respectively (Fig. \ref{results_fig_1} B). We further confirmed that NALSM's increased performance resulted from the improved computational properties of its liquid by measuring each model's liquid kernel quality. This encompassed both the linear separation and generalization capability of the liquid (See \ref{sup_kernel_quality}). Higher model accuracy corresponded to higher kernel quality for all $4$ models (Fig. \ref{results_fig_2_acc_kq} B). This is a further indication that near-critical dynamics and astrocyte-modulated STDP contributed to the NALSM's performance increase.

\subsection{NALSM maintained performance with sparse neuron-astrocyte connectivity}
The NALSM maintained its accuracy advantage even with neuron-astrocyte connection densities as low as $10\%$ (Fig. \ref{results_fig_3_ABLATION}). In the brain, astrocytes contact only approximately $65\%$ of all synapses in their surroundings \cite{Zhou2019}. We tested NALSM performance as a function of neuron-astrocyte connection density (See \ref{methods_experiments}). The NALSM mean accuracy decreased marginally with increasingly sparse connectivity, while variability in performance was minimal across densities. At 10\% connectivity, average NALSM accuracy decreased by $0.36\%$ for MNIST and $0.14\%$ for N-MNIST compared to $100\%$ connection density. In both cases, average NALSM accuracy was still above LSM and LSM+APSTDP average accuracy.

\subsection{Larger liquids increased NALSM accuracy}
The NALSM accuracy improved with increased liquid size, saturating at approximately $8,000$ neurons (See Appendix \ref{appendix_lsm_sizes}). NALSM$8000$ achieved a top accuracy of $97.61\%$ ($97.49 \pm 0.11\%$) on MNIST, $97.51\%$ ($97.42 \pm 0.07\%$) on N-MNIST, and $85.84\%$ ($85.61 \pm 0.18\%$) on Fashion-MNIST. Compared to previously reported benchmarks on MNIST and Fashion-MNIST, the NALSM$8000$ outperformed all brain-inspired learning methods that do not use backpropagation of gradients or its approximation through feedback alignment \cite{Zhao2020GLSNNAM,Samadi2017DeepLW}, with the exception of \cite{Zhang_2018_balsnn} for MNIST. While \cite{Zhang_2018_balsnn} demonstrated that a fully-connected 2-layer spiking network can achieve high accuracy through a combination of biologically-plausible plasticity rules, it is not clear how such an approach would scale to more layers without some form of backpropagation. Conversely, multi-layered LSMs have been shown to work without backpropagation \cite{soures_2019,wang_2016}. Further, NALSM$8000$ used approximately $1/3$ ($\approx  1,199,407 \pm 453$) of number of trainable(plastic) connections as in \cite{Zhang_2018_balsnn}. Compared to top accuracies reported for fully-connected multi-layered spiking neural networks trained with backpropagation, the NALSM$8000$ achieved comparable performance on all datasets; outperforming multiple reported results on MNIST and N-MNIST (Table \ref{acc_table}) (See Appendix \ref{appendix_number_plastic_parameters}).

\begin{table}
\caption{Comparison to brain-inspired and fully-connected multi-layer spiking neural networks.}
\centering
\begin{tabular}{lccc@{}}
\toprule
          \textbf{Model} & \textbf{Layers} & \textbf{Learning Method}  & \textbf{Accuracy} \\ \midrule
 \multicolumn{4}{@{}l}{\textbf{Dataset: MNIST}}\\
     Unsupervised-SNN \cite{Diehl_2015} & 2 & STDP & $95\%$      \\
     Multi-liquid LSM \cite{multi_liquid_lsm_2019} & 2 & GD on last layer & $95.5\%$      \\
     {\bf{NALSM1000}} & {\bf{2}} & {\bf{astro-STDP, GD on last layer}} & $\textbf{96.15\%}$ \\
    LIF-BA \cite{Samadi2017DeepLW} & 3 & Broadcast feedback alignment & $97.09\%$ \\
     Temporal SNN \cite{Mostafa2018SupervisedLB}& 2 & Temporal backpropagation & $97.2\%$ \\
     STiDi-BP \cite{Mirsadeghi_2021}& 2 & Backpropagation & $97.4\%$ \\
    {\bf{NALSM8000}} & {\bf{2}} & {\bf{astro-STDP, GD on last layer}} & $\textbf{97.61\%}$ \\
    SN \cite{OConnor2016DeepSN} & 3 & Backpropagation & $97.93\%$      \\
    GLSNN \cite{Zhao2020GLSNNAM} & 4 & Global feedback alignment, STDP & $98.62\%$     \\
    Balance-SNN \cite{Zhang_2018_balsnn} & 2 & Equi-prop, STDP, STP & $98.64\%$ \\
    BPSNN \cite{Lee_2016_backpropsnn}  & 3 & Backpropagation & $98.88\%$      \\
    STBP \cite{Wu_2018_ssnbackprop} & 2 & Spatial and temporal backpropagation     & $98.89\%$      \\
  \midrule
     \multicolumn{4}{@{}l}{\textbf{Dataset: N-MNIST}}\\
    DECOLLE \cite{Kaiser_2020} & 3 & Backpropagation      & $96\%$   \\
    {\bf{NALSM1000}} & {\bf{2}} & {\bf{astro-STDP, GD on last layer}} & $\textbf{96.13\%}$ \\
    AER-SNN \cite{Liu2020EffectiveAO} & 2 & Backpropagation & $96.3\%$    \\
    {\bf{NALSM8000}} & {\bf{2}} & {\bf{astro-STDP, GD on last layer}} & $\textbf{97.51\%}$ \\
    BPSNN \cite{Lee_2016_backpropsnn} & 3 & Backpropagation & $98.74\%$ \\
    STBP \cite{Wu_2018_ssnbackprop} & 2 & Spatial and temporal backpropagation     & $98.78\%$      \\
    SLAYER \cite{Shrestha2018SLAYERSL} & 3 & Backpropagation & $98.89\%$   \\
  \midrule
     \multicolumn{4}{@{}l}{\textbf{Dataset: Fashion-MNIST}}\\
    VPSNN \cite{Zhang_2018} & 2 & Equi-prop, STDP & $82.69\%$ \\
    {\bf{NALSM1000}} & {\bf{2}} & {\bf{astro-STDP, GD on last layer}} & $\textbf{83.54\%}$ \\
    Unsupervised-SNN \cite{HAO2020} & 2 & STDP & $85.31\%$ \\
    {\bf{NALSM8000}} & {\bf{2}} & {\bf{astro-STDP, GD on last layer}} & $\textbf{85.84\%}$ \\
    BS4NN \cite{Kheradpisheh_2007} & 2 & Temporal backpropagation & $87.3\%$ \\
    GLSNN \cite{Zhao2020GLSNNAM} & 4 & Global feedback alignment, STDP & $89.05\%$ \\
  \bottomrule
  *GD: gradient descent
\end{tabular}
\label{acc_table}
\end{table}

\section{Discussion and Broader Impact}
Ironically, LSMs are one of the most brain-like and at the same time one of the most difficult to train learning models. Here, we proposed an astrocyte model that merged critical branching dynamics and STDP into a single liquid, thereby simultaneously improving LSM performance and decreasing data-specific tuning. We showed that the synergy of STDP and near-critical branching dynamics improved the computational capacity of the liquid, which translated to better than state-of-the-art LSM accuracy on MNIST and N-MNIST, and do so with minimal added computational cost (See Appendix \ref{appendix_computational_cost}). Our results indicate that, given a large enough liquid, NALSM performance compares to current fully-connected multi-layer spiking neural networks trained via backpropagation. 

The reported narrowing of the performance gap between brain-inspired LSM and deep networks suggests that studying the interaction among the brain's computational principles can help our learning models to reach human-like performance. Indeed, our results demonstrate that the synergy of brain-inspired astrocyte-modulated STDP and near-critical dynamics resulted in the superior performance of NALSM compared to 1) a LSM with critical dynamics but without STDP, and 2) a LSM with STDP, but without critical dynamics. Aligning with other studies showing that liquid topology impacts LSM accuracy \cite{JU201339}, we also showed that a brain-inspired, sparse, 3D-distance-based network architecture can improve the computational capacity of a single liquid. Specifically, our baseline 3D LSM achieved comparable accuracy to the multi-liquid LSM \cite{multi_liquid_lsm_2019}, which improved performance of a single dimensionless liquid by partitioning it into multiple liquids. While we demonstrated NALSM performance using only the 3D-distance-based network architecture, our proposed astrocyte modulation method does not depend on network topology and, therefore, is applicable to other types of topology. In fact, our approach is also extendable to multi-liquid architectures and other local plasticity rules that follow STDP's separation of potentiation and depression components. 

The astrocyte-modulated LSM learning framework is also compatible with the emerging neuromorphic hardware. This is because the gradient descent that we used for training the linear output can be replaced by a single-layer spike based learning rule \cite{Ponulak_2010,Urbanczik_2009,Gutig_2006}. This makes NALSM compatible with neuromorphic hardware, exploiting in full its advantages \cite{Tang_2019_2}. For example, our method can leverage even further the energy efficiency of neuromorphic chips, by virtue of its low spiking rates. In line with biological ranges \cite{Roxin2011}, NALSM had spiking rates that ranged from $12$ $Hz$ to $37$ $Hz$, depending on the input sample. These rates can be reduced further, by modifying input encoding, since liquid spiking rates are directly adjusted by the astrocyte based on input spiking rates (Fig. \ref{methods_fig_0}).

Here, we demonstrated a possible connection between the near-critical branching dynamics of the NALSM liquid and the edge-of-chaos transition (See Appendix \ref{appendix_edge_of_chaos}). The critical branching transition has been extensively used to model critical dynamics in brain networks \cite{Shriki2013,Beggs2003}. Focusing on the computational benefits of criticality, machine learning has mostly examined network dynamics at the edge-of-chaos transition. Although the presence of one transition does not guarantee the existence of the other, both transitions are well connected to the same result, an improved computational performance \cite{Haldeman_2005}. Indeed, the computational performance of systems poised at a critical phase transition has been widely studied both experimentally \cite{Shew2011} and theoretically \cite{de_Arcangelis2010}, and are well-connected to both edge-of-chaos \cite{Legenstein2007,Bertschinger2004} and critical branching transitions \cite{Haldeman_2005,balafrej2020pcritical}. Networks operating at near-criticality are believed to have simultaneous access to the computational properties (learning and memory) of both phases, which results in 1) maximizing their information processing capacity \cite{Shew2011}, 2) optimizing their dynamical range \cite{Shew2009,Kinouchi2006}, and 3) expanding their number of metastable states \cite{Haldeman_2005}. Hence, it is not surprising that the NALSM's astrocyte imposed near-critical branching dynamics resulted in improved accuracy and generalization capabilities as observed in LSMs with edge-of-chaos dynamics \cite{Legenstein2007,Bertschinger2004}, while adding the benefit of a neuromorphic compatibility and self-organized criticality.

Our work shows how insights from modern cellular neuroscience can synergize with neuromorphic computing, and lead to novel intelligent systems, spurring the dialogue between artificial intelligence and brain sciences. Indeed, given that the known neuronal mechanisms are too slow and uncoordinated in the brain to modulate STDP \cite{Zenke2017, Zenke2013, Watt2010}, it is an open question how neurons modulate synaptic plasticity. Our demonstration that the distinct temporal and spatial mechanisms of astrocytes may modulate STDP and subsequently regulate network dynamics, questions the neuron as the only processing unit in the brain \cite{RN55,RN54,RN53}. In that sense, it helps in dismantling the 100-year old dogma that ``brain = neurons'', and tackle the absence of astrocytes in both prevailing computational hypotheses on how the brain learns and efforts to translate such knowledge to effective models of intelligence.

By showing how astrocyte-modulated STDP can maximize computational performance near criticality, we aimed to broaden the applicability of the LSM to complex spatio-temporal problems that require integration of data over multiple sources and time-scales, thereby, making LSMs suitable for real-life applications of edge computing. Our so far results suggest that this is a direction worth pursuing.

\section*{Acknowledgements}
This work is supported by the National Center for Medical Rehabilitation Research (NIH/NICHD) $K12HD093427$ Grant and by the Rutgers Office of Research and Innovation. Any findings, conclusions, and opinions expressed in this material are those of the authors and do not necessarily reflect the views of the NIH or Rutgers University.

\bibliographystyle{unsrtnat}

\bibliography{bib}

\begin{thebibliography}{9}
\providecommand{\natexlab}[1]{#1}
\providecommand{\url}[1]{\texttt{#1}}
\expandafter\ifx\csname urlstyle\endcsname\relax
  \providecommand{\doi}[1]{doi: #1}\else
  \providecommand{\doi}{doi: \begingroup \urlstyle{rm}\Url}\fi

\bibitem[Lecun et~al.(1998)Lecun, Bottou, Bengio, and Haffner]{mnist_bib_0}
Y.~Lecun, L.~Bottou, Y.~Bengio, and P.~Haffner.
\newblock Gradient-based learning applied to document recognition.
\newblock \emph{Proceedings of the IEEE}, 86\penalty0 (11):\penalty0
  2278--2324, 1998.
\newblock \doi{10.1109/5.726791}.

\bibitem[Orchard et~al.(2015)Orchard, Jayawant, Cohen, and
  Thakor]{nmnist_bib_0}
Garrick Orchard, Ajinkya Jayawant, Gregory~K. Cohen, and Nitish Thakor.
\newblock Converting static image datasets to spiking neuromorphic datasets
  using saccades.
\newblock \emph{Frontiers in Neuroscience}, 9:\penalty0 437, 2015.
\newblock ISSN 1662-453X.
\newblock \doi{10.3389/fnins.2015.00437}.

\bibitem[Xiao et~al.(2017)Xiao, Rasul, and Vollgraf]{fmnist_bib_0}
Han Xiao, Kashif Rasul, and Roland Vollgraf.
\newblock Fashion-mnist: a novel image dataset for benchmarking machine
  learning algorithms, 2017.

\bibitem[Jin and Li(2016)]{apstdp_bib_0}
Yingyezhe Jin and Peng Li.
\newblock Ap-stdp: A novel self-organizing mechanism for efficient reservoir
  computing.
\newblock In \emph{2016 International Joint Conference on Neural Networks
  (IJCNN)}, pages 1158--1165, 2016.
\newblock \doi{10.1109/IJCNN.2016.7727328}.

\bibitem[Ku\ifmmode~\acute{s}\else \'{s}\fi{}mierz
  et~al.(2020)Ku\ifmmode~\acute{s}\else \'{s}\fi{}mierz, Ogawa, and
  Toyoizumi]{kus_2020_0}
\L{}ukasz Ku\ifmmode~\acute{s}\else \'{s}\fi{}mierz, Shun Ogawa, and Taro
  Toyoizumi.
\newblock Edge of chaos and avalanches in neural networks with heavy-tailed
  synaptic weight distribution.
\newblock \emph{Phys. Rev. Lett.}, 125:\penalty0 028101, Jul 2020.
\newblock \doi{10.1103/PhysRevLett.125.028101}.

\bibitem[van Vreeswijk and Sompolinsky(1996)]{Vreeswijk_1996_0}
C.~van Vreeswijk and H.~Sompolinsky.
\newblock Chaos in neuronal networks with balanced excitatory and inhibitory
  activity.
\newblock \emph{Science}, 274\penalty0 (5293):\penalty0 1724--1726, 1996.
\newblock \doi{10.1126/science.274.5293.1724}.

\bibitem[Krauss et~al.(2019)Krauss, Schuster, Dietrich, Schilling, Schulze, and
  Metzner]{Krauss_2019_0}
Patrick Krauss, Marc Schuster, Verena Dietrich, Achim Schilling, Holger
  Schulze, and Claus Metzner.
\newblock Weight statistics controls dynamics in recurrent neural networks.
\newblock \emph{PLOS ONE}, 14:\penalty0 1--13, 04 2019.
\newblock \doi{10.1371/journal.pone.0214541}.

\bibitem[Ostojic(2014)]{Ostojic_2014_0}
Srdjan Ostojic.
\newblock Two types of asynchronous activity in networks of excitatory and
  inhibitory spiking neurons.
\newblock \emph{Nature Neuroscience}, 17:\penalty0 594--600, 2014.
\newblock ISSN 1546-1726.
\newblock \doi{10.1038/nn.3658}.

\bibitem[Rajan et~al.(2010)Rajan, Abbott, and Sompolinsky]{Rajan_2010_0}
Kanaka Rajan, L.~F. Abbott, and Haim Sompolinsky.
\newblock Stimulus-dependent suppression of chaos in recurrent neural networks.
\newblock \emph{Phys. Rev. E}, 82:\penalty0 011903, Jul 2010.
\newblock \doi{10.1103/PhysRevE.82.011903}.

\end{thebibliography}


\begin{thebibliography}{97}
\providecommand{\natexlab}[1]{#1}
\providecommand{\url}[1]{\texttt{#1}}
\expandafter\ifx\csname urlstyle\endcsname\relax
  \providecommand{\doi}[1]{doi: #1}\else
  \providecommand{\doi}{doi: \begingroup \urlstyle{rm}\Url}\fi

\bibitem[Kendall and Kumar(2020)]{Kendall_2020}
Jack~D. Kendall and Suhas Kumar.
\newblock The building blocks of a brain-inspired computer.
\newblock \emph{Applied Physics Reviews}, 7\penalty0 (1):\penalty0 011305,
  2020.
\newblock \doi{10.1063/1.5129306}.

\bibitem[Tang et~al.(2020)Tang, Kumar, and Michmizos]{Tang_2020}
Guangzhi Tang, Neelesh Kumar, and Konstantinos~P. Michmizos.
\newblock Reinforcement co-learning of deep and spiking neural networks for
  energy-efficient mapless navigation with neuromorphic hardware.
\newblock In \emph{2020 IEEE/RSJ International Conference on Intelligent Robots
  and Systems (IROS)}, pages 6090--6097, 2020.
\newblock \doi{10.1109/IROS45743.2020.9340948}.

\bibitem[Davies et~al.(2018)Davies, Srinivasa, Lin, Chinya, Cao, Choday, Dimou,
  Joshi, Imam, Jain, Liao, Lin, Lines, Liu, Mathaikutty, McCoy, Paul, Tse,
  Venkataramanan, Weng, Wild, Yang, and Wang]{lif_loihi}
Mike Davies, Narayan Srinivasa, Tsung-Han Lin, Gautham Chinya, Yongqiang Cao,
  Sri~Harsha Choday, Georgios Dimou, Prasad Joshi, Nabil Imam, Shweta Jain,
  Yuyun Liao, Chit-Kwan Lin, Andrew Lines, Ruokun Liu, Deepak Mathaikutty,
  Steven McCoy, Arnab Paul, Jonathan Tse, Guruguhanathan Venkataramanan,
  Yi-Hsin Weng, Andreas Wild, Yoonseok Yang, and Hong Wang.
\newblock Loihi: A neuromorphic manycore processor with on-chip learning.
\newblock \emph{IEEE Micro}, 38\penalty0 (1):\penalty0 82--99, 2018.
\newblock \doi{10.1109/MM.2018.112130359}.

\bibitem[Tang et~al.(2019{\natexlab{a}})Tang, Shah, and Michmizos]{Tang_2019}
Guangzhi Tang, Arpit Shah, and Konstantinos~P. Michmizos.
\newblock Spiking neural network on neuromorphic hardware for energy-efficient
  unidimensional slam.
\newblock In \emph{2019 IEEE/RSJ International Conference on Intelligent Robots
  and Systems (IROS)}, pages 4176--4181, 2019{\natexlab{a}}.
\newblock \doi{10.1109/IROS40897.2019.8967864}.

\bibitem[Cao et~al.(2020)Cao, Liu, Meng, and Sun]{Cao_2020}
Keyan Cao, Yefan Liu, Gongjie Meng, and Qimeng Sun.
\newblock An overview on edge computing research.
\newblock \emph{IEEE Access}, 8:\penalty0 85714--85728, 2020.
\newblock \doi{10.1109/ACCESS.2020.2991734}.

\bibitem[Chen and Ran(2019)]{Chen_2019}
Jiasi Chen and Xukan Ran.
\newblock Deep learning with edge computing: A review.
\newblock \emph{Proceedings of the IEEE}, 107\penalty0 (8):\penalty0
  1655--1674, 2019.
\newblock \doi{10.1109/JPROC.2019.2921977}.

\bibitem[Maass et~al.(2002)Maass, Natschläger, and Markram]{Maass2002}
Wolfgang Maass, Thomas Natschläger, and Henry Markram.
\newblock Real-time computing without stable states: A new framework for neural
  computation based on perturbations.
\newblock \emph{Neural Computation}, 14\penalty0 (11):\penalty0 2531--2560,
  2002.
\newblock \doi{10.1162/089976602760407955}.

\bibitem[Soures~Nicholas(2019)]{soures_2019}
Kudithipudi~Dhireesha Soures~Nicholas.
\newblock Deep liquid state machines with neural plasticity for video activity
  recognition.
\newblock \emph{Frontiers in Neuroscience}, 13:\penalty0 686, 2019.
\newblock \doi{10.3389/fnins.2019.00686}.

\bibitem[Ponghiran et~al.(2019)Ponghiran, Srinivasan, and Roy]{Ponghiran2019}
Wachirawit Ponghiran, Gopalakrishnan Srinivasan, and Kaushik Roy.
\newblock Reinforcement learning with low-complexity liquid state machines.
\newblock \emph{Frontiers in Neuroscience}, 13:\penalty0 883, 2019.
\newblock ISSN 1662-453X.
\newblock \doi{10.3389/fnins.2019.00883}.

\bibitem[Wang and Li(2016)]{wang_2016}
Qian Wang and Peng Li.
\newblock D-lsm: Deep liquid state machine with unsupervised recurrent
  reservoir tuning.
\newblock In \emph{2016 23rd International Conference on Pattern Recognition
  (ICPR)}, pages 2652--2657, 2016.
\newblock \doi{10.1109/ICPR.2016.7900035}.

\bibitem[Liu et~al.(2020{\natexlab{a}})Liu, Liu, and Yi]{Shiya_2020}
Shiya Liu, Lingjia Liu, and Yang Yi.
\newblock Quantized reservoir computing on edge devices for communication
  applications.
\newblock In \emph{2020 IEEE/ACM Symposium on Edge Computing (SEC)}, pages
  445--449, 2020{\natexlab{a}}.
\newblock \doi{10.1109/SEC50012.2020.00068}.

\bibitem[Li et~al.(2020)Li, Wang, Wang, and Xu]{Li_2020}
Shiming Li, Lei Wang, Shiying Wang, and Weixia Xu.
\newblock Liquid state machine applications mapping for noc-based neuromorphic
  platforms.
\newblock In Dezun Dong, Xiaoli Gong, Cunlu Li, Dongsheng Li, and Junjie Wu,
  editors, \emph{Advanced Computer Architecture}, pages 277--289, Singapore,
  2020. Springer Singapore.
\newblock ISBN 978-981-15-8135-9.

\bibitem[Ku et~al.(2018)Ku, Liu, Jin, Samal, Li, and Lim]{Ku_2018}
Bon~Woong Ku, Yu~Liu, Yingyezhe Jin, Sandeep Samal, Peng Li, and Sung~Kyu Lim.
\newblock Design and architectural co-optimization of monolithic 3d liquid
  state machine-based neuromorphic processor.
\newblock In \emph{2018 55th ACM/ESDA/IEEE Design Automation Conference (DAC)},
  pages 1--6, 2018.
\newblock \doi{10.1109/DAC.2018.8465837}.

\bibitem[Rosselló et~al.(2016)Rosselló, Alomar, Morro, Oliver, and
  Canals]{Rossello_2016}
Josep~L. Rosselló, Miquel~L. Alomar, Antoni Morro, Antoni Oliver, and Vincent
  Canals.
\newblock High-density liquid-state machine circuitry for time-series
  forecasting.
\newblock \emph{International Journal of Neural Systems}, 26\penalty0
  (05):\penalty0 1550036, 2016.
\newblock \doi{10.1142/S0129065715500367}.
\newblock PMID: 26906454.

\bibitem[Balafrej and Rouat(2020)]{balafrej2020pcritical}
Ismael Balafrej and Jean Rouat.
\newblock P-critical: A reservoir autoregulation plasticity rule for
  neuromorphic hardware, 2020.

\bibitem[Brodeur and Rouat(2012)]{Brodeur_2012}
Simon Brodeur and Jean Rouat.
\newblock Regulation toward self-organized criticality in a recurrent spiking
  neural reservoir.
\newblock In Alessandro E.~P. Villa, W{\l}odzis{\l}aw Duch, P{\'e}ter {\'E}rdi,
  Francesco Masulli, and G{\"u}nther Palm, editors, \emph{Artificial Neural
  Networks and Machine Learning -- ICANN 2012}, pages 547--554, Berlin,
  Heidelberg, 2012. Springer Berlin Heidelberg.

\bibitem[Jin and Li(2016)]{apstdp_bib}
Yingyezhe Jin and Peng Li.
\newblock Ap-stdp: A novel self-organizing mechanism for efficient reservoir
  computing.
\newblock In \emph{2016 International Joint Conference on Neural Networks
  (IJCNN)}, pages 1158--1165, 2016.
\newblock \doi{10.1109/IJCNN.2016.7727328}.

\bibitem[Norton and Ventura(2006)]{Norton_2006}
D.~Norton and D.~Ventura.
\newblock Preparing more effective liquid state machines using hebbian
  learning.
\newblock In \emph{The 2006 IEEE International Joint Conference on Neural
  Network Proceedings}, pages 4243--4248, 2006.
\newblock \doi{10.1109/IJCNN.2006.246996}.

\bibitem[Legenstein and Maass(2007)]{Legenstein2007}
Robert Legenstein and Wolfgang Maass.
\newblock Edge of chaos and prediction of computational performance for neural
  circuit models.
\newblock \emph{Neural Networks}, 20\penalty0 (3):\penalty0 323 -- 334, 2007.
\newblock ISSN 0893-6080.
\newblock \doi{10.1016/j.neunet.2007.04.017}.
\newblock Echo State Networks and Liquid State Machines.

\bibitem[Bertschinger and Natschläger(2004)]{Bertschinger2004}
Nils Bertschinger and Thomas Natschläger.
\newblock Real-time computation at the edge of chaos in recurrent neural
  networks.
\newblock \emph{Neural Computation}, 16\penalty0 (7):\penalty0 1413--1436,
  2004.
\newblock \doi{10.1162/089976604323057443}.

\bibitem[Langton(1990)]{Langton1990}
Chris~G. Langton.
\newblock Computation at the edge of chaos: phase transitions and emergent
  computation.
\newblock \emph{Physica D: Nonlinear Phenomena}, 42\penalty0 (1):\penalty0 12
  -- 37, 1990.
\newblock ISSN 0167-2789.
\newblock \doi{10.1016/0167-2789(90)90064-V}.

\bibitem[Shew et~al.(2011)Shew, Yang, Yu, Roy, and Plenz]{Shew2011}
Woodrow~L. Shew, Hongdian Yang, Shan Yu, Rajarshi Roy, and Dietmar Plenz.
\newblock Information capacity and transmission are maximized in balanced
  cortical networks with neuronal avalanches.
\newblock \emph{Journal of Neuroscience}, 31\penalty0 (1):\penalty0 55--63,
  2011.
\newblock ISSN 0270-6474.
\newblock \doi{10.1523/JNEUROSCI.4637-10.2011}.

\bibitem[de~Arcangelis and Herrmann(2010)]{de_Arcangelis2010}
Lucilla de~Arcangelis and Hans~J. Herrmann.
\newblock Learning as a phenomenon occurring in a critical state.
\newblock \emph{Proceedings of the National Academy of Sciences}, 107\penalty0
  (9):\penalty0 3977--3981, 2010.
\newblock ISSN 0027-8424.
\newblock \doi{10.1073/pnas.0912289107}.

\bibitem[Kinouchi and Copelli(2006)]{Kinouchi2006}
Osame Kinouchi and Mauro Copelli.
\newblock Optimal dynamical range of excitable networks at criticality.
\newblock \emph{Nature Physics}, 2\penalty0 (5):\penalty0 348--351, 2006.
\newblock \doi{10.1038/nphys289}.

\bibitem[Haldeman and Beggs(2005)]{Haldeman_2005}
Clayton Haldeman and John~M. Beggs.
\newblock Critical branching captures activity in living neural networks and
  maximizes the number of metastable states.
\newblock \emph{Phys. Rev. Lett.}, 94:\penalty0 058101, Feb 2005.
\newblock \doi{10.1103/PhysRevLett.94.058101}.

\bibitem[Beggs and Plenz(2003)]{Beggs2003}
John~M. Beggs and Dietmar Plenz.
\newblock Neuronal avalanches in neocortical circuits.
\newblock \emph{Journal of Neuroscience}, 23\penalty0 (35):\penalty0
  11167--11177, 2003.
\newblock ISSN 0270-6474.
\newblock \doi{10.1523/JNEUROSCI.23-35-11167.2003}.

\bibitem[Shriki et~al.(2013)Shriki, Alstott, Carver, Holroyd, Henson, Smith,
  Coppola, Bullmore, and Plenz]{Shriki2013}
Oren Shriki, Jeff Alstott, Frederick Carver, Tom Holroyd, Richard~N.A. Henson,
  Marie~L. Smith, Richard Coppola, Edward Bullmore, and Dietmar Plenz.
\newblock Neuronal avalanches in the resting meg of the human brain.
\newblock \emph{Journal of Neuroscience}, 33\penalty0 (16):\penalty0
  7079--7090, 2013.
\newblock ISSN 0270-6474.
\newblock \doi{10.1523/JNEUROSCI.4286-12.2013}.

\bibitem[Chialvo(2010)]{Chialvo2010}
Dante~R. Chialvo.
\newblock Emergent complex neural dynamics.
\newblock \emph{Nature Physics}, 6:\penalty0 744--750, 2010.
\newblock ISSN 1745-2481.
\newblock \doi{10.1038/nphys1803}.

\bibitem[Feldman(2009)]{Feldman2009}
Daniel~E. Feldman.
\newblock Synaptic mechanisms for plasticity in neocortex.
\newblock \emph{Annual Review of Neuroscience}, 32\penalty0 (1):\penalty0
  33--55, 2009.
\newblock \doi{10.1146/annurev.neuro.051508.135516}.
\newblock PMID: 19400721.

\bibitem[Stepp et~al.(2015)Stepp, Plenz, and Srinivasa]{stepp2015}
Nigel Stepp, Dietmar Plenz, and Narayan Srinivasa.
\newblock Synaptic plasticity enables adaptive self-tuning critical networks.
\newblock \emph{PLOS Computational Biology}, 11\penalty0 (1):\penalty0 1--28,
  01 2015.
\newblock \doi{10.1371/journal.pcbi.1004043}.

\bibitem[Zenke and Gerstner(2017)]{Zenke2017}
Friedemann Zenke and Wulfram Gerstner.
\newblock Hebbian plasticity requires compensatory processes on multiple
  timescales.
\newblock \emph{Philosophical Transactions of the Royal Society B: Biological
  Sciences}, 372\penalty0 (1715):\penalty0 20160259, 2017.
\newblock \doi{10.1098/rstb.2016.0259}.

\bibitem[Watt and Desai(2010)]{Watt2010}
Alanna Watt and Niraj Desai.
\newblock Homeostatic plasticity and stdp: keeping a neuron's cool in a
  fluctuating world.
\newblock \emph{Frontiers in Synaptic Neuroscience}, 2:\penalty0 5, 2010.
\newblock ISSN 1663-3563.
\newblock \doi{10.3389/fnsyn.2010.00005}.

\bibitem[Abbott and Nelson(2000)]{Abbott2000}
L.~F. Abbott and Sacha~B. Nelson.
\newblock Synaptic plasticity: taming the beast.
\newblock \emph{Nature Neuroscience}, 3\penalty0 (11):\penalty0 1178--1183,
  2000.
\newblock ISSN 1546-1726.
\newblock \doi{10.1038/81453}.

\bibitem[Perea and Araque(2010)]{RN46}
Gertrudis Perea and Alfonso Araque.
\newblock Glia modulates synaptic transmission.
\newblock \emph{Brain Research Reviews}, 63\penalty0 (1):\penalty0 93--102,
  2010.
\newblock ISSN 0165-0173.

\bibitem[Han et~al.(2013)Han, Chen, Wang, Windrem, Wang, Shanz, Xu, Oberheim,
  Bekar, and Betstadt]{Han2013}
Xiaoning Han, Michael Chen, Fushun Wang, Martha Windrem, Su~Wang, Steven Shanz,
  Qiwu Xu, Nancy~Ann Oberheim, Lane Bekar, and Sarah Betstadt.
\newblock Forebrain engraftment by human glial progenitor cells enhances
  synaptic plasticity and learning in adult mice.
\newblock \emph{Cell Stem Cell}, 12\penalty0 (3):\penalty0 342--353, 2013.
\newblock ISSN 1934-5909.

\bibitem[Tewari and Parpura(2013)]{Tewari2013}
Shivendra Tewari and Vladimir Parpura.
\newblock A possible role of astrocytes in contextual memory retrieval: An
  analysis obtained using a quantitative framework.
\newblock \emph{Frontiers in Computational Neuroscience}, 7:\penalty0 145,
  2013.
\newblock ISSN 1662-5188.
\newblock \doi{10.3389/fncom.2013.00145}.

\bibitem[Navarrete and Araque(2010)]{RN52}
Marta Navarrete and Alfonso Araque.
\newblock Endocannabinoids potentiate synaptic transmission through stimulation
  of astrocytes.
\newblock \emph{Neuron}, 68\penalty0 (1):\penalty0 113--126, 2010.
\newblock ISSN 0896-6273.

\bibitem[Adamsky et~al.(2018)Adamsky, Kol, Kreisel, Doron, Ozeri-Engelhard,
  Melcer, Refaeli, Horn, Regev, and Groysman]{RN50}
Adar Adamsky, Adi Kol, Tirzah Kreisel, Adi Doron, Nofar Ozeri-Engelhard, Talia
  Melcer, Ron Refaeli, Henrike Horn, Limor Regev, and Maya Groysman.
\newblock Astrocytic activation generates de novo neuronal potentiation and
  memory enhancement.
\newblock \emph{Cell}, 174\penalty0 (1):\penalty0 59--71. e14, 2018.
\newblock ISSN 0092-8674.

\bibitem[Petrelli et~al.(2020)Petrelli, Dallérac, Pucci, Calì, Zehnder,
  Sultan, Lecca, Chicca, Ivanov, Asensio, Gundersen, Toni, Knott, Magara,
  Gertsch, Kirchhoff, Déglon, Giros, Edwards, Mothet, and
  Bezzi]{Petrelli_2020}
Francesco Petrelli, Glenn Dallérac, Luca Pucci, Corrado Calì, Tamara Zehnder,
  Sébastien Sultan, Salvatore Lecca, Andrea Chicca, Andrei Ivanov, Cédric~S.
  Asensio, Vidar Gundersen, Nicolas Toni, Graham~William Knott, Fulvio Magara,
  Jürg Gertsch, Frank Kirchhoff, Nicole Déglon, Bruno Giros, Robert~H.
  Edwards, Jean-Pierre Mothet, and Paola Bezzi.
\newblock Dysfunction of homeostatic control of dopamine by astrocytes in the
  developing prefrontal cortex leads to cognitive impairments.
\newblock \emph{Molecular Psychiatry}, 25\penalty0 (4):\penalty0 732--749,
  2020.
\newblock \doi{10.1038/s41380-018-0226-y}.

\bibitem[Manninen et~al.(2020)Manninen, Saudargiene, and Linne]{Manninen_2020}
Tiina Manninen, Ausra Saudargiene, and Marja-Leena Linne.
\newblock Astrocyte-mediated spike-timing-dependent long-term depression
  modulates synaptic properties in the developing cortex.
\newblock \emph{PLOS Computational Biology}, 16\penalty0 (11):\penalty0 1--29,
  11 2020.
\newblock \doi{10.1371/journal.pcbi.1008360}.

\bibitem[Sibille et~al.(2014)Sibille, Pannasch, and Rouach]{Sibille_2014}
Jérémie Sibille, Ulrike Pannasch, and Nathalie Rouach.
\newblock Astroglial potassium clearance contributes to short-term plasticity
  of synaptically evoked currents at the tripartite synapse.
\newblock \emph{The Journal of Physiology}, 592\penalty0 (1):\penalty0 87--102,
  2014.
\newblock \doi{10.1113/jphysiol.2013.261735}.

\bibitem[Min and Nevian(2012)]{Min_2012}
Rogier Min and Thomas Nevian.
\newblock Astrocyte signaling controls spike timing–dependent depression at
  neocortical synapses.
\newblock \emph{Nature Neuroscience}, 15\penalty0 (5):\penalty0 746--753, 2012.
\newblock \doi{10.1038/nn.3075}.

\bibitem[Thrane et~al.(2012)Thrane, Rangroo~Thrane, Zeppenfeld, Lou, Xu,
  Nagelhus, and Nedergaard]{Thrane2012}
Alexander~Stanley Thrane, Vinita Rangroo~Thrane, Douglas Zeppenfeld, Nanhong
  Lou, Qiwu Xu, Erlend~Arnulf Nagelhus, and Maiken Nedergaard.
\newblock General anesthesia selectively disrupts astrocyte calcium signaling
  in the awake mouse cortex.
\newblock \emph{Proceedings of the National Academy of Sciences}, 109\penalty0
  (46):\penalty0 18974--18979, 2012.
\newblock ISSN 0027-8424.
\newblock \doi{10.1073/pnas.1209448109}.

\bibitem[Foley et~al.(2017)Foley, Blutstein, Lee, Erneux, Halassa, and
  Haydon]{Foley2017}
Jeannine Foley, Tamara Blutstein, SoYoung Lee, Christophe Erneux, Michael~M.
  Halassa, and Philip Haydon.
\newblock Astrocytic ip3/ca2+ signaling modulates theta rhythm and rem sleep.
\newblock \emph{Frontiers in Neural Circuits}, 11:\penalty0 3, 2017.
\newblock ISSN 1662-5110.
\newblock \doi{10.3389/fncir.2017.00003}.

\bibitem[Bojarskaite et~al.(2020)Bojarskaite, Bj{\o}rnstad, Pettersen, Cunen,
  Hermansen, {\r A}bj{\o}rsbr{\r a}ten, Sprengel, Vervaeke, Tang, Enger, and
  Nagelhus]{Bojarskaite2019}
Laura Bojarskaite, Daniel~M. Bj{\o}rnstad, Klas~H. Pettersen, C{\'e}line Cunen,
  Gudmund~Horn Hermansen, Knut~Sindre {\r A}bj{\o}rsbr{\r a}ten, Rolf Sprengel,
  Koen Vervaeke, Wannan Tang, Rune Enger, and Erlend~A. Nagelhus.
\newblock Astrocytic ca2+ signaling is reduced during sleep and is involved in
  the regulation of slow wave sleep.
\newblock \emph{Nature Communications}, 11:\penalty0 3240, 2020.
\newblock ISSN 2041-1723.
\newblock \doi{10.1038/s41467-020-17062-2}.

\bibitem[Ingiosi et~al.(2020)Ingiosi, Hayworth, Harvey, Singletary, Rempe,
  Wisor, and Frank]{Ingiosi2019}
Ashley~M. Ingiosi, Christopher~R. Hayworth, Daniel~O. Harvey, Kristan~G.
  Singletary, Michael~J. Rempe, Jonathan~P. Wisor, and Marcos~G. Frank.
\newblock A role for astroglial calcium in mammalian sleep and sleep
  regulation.
\newblock \emph{Current Biology}, 30:\penalty0 4373--4383.e7, 2020.
\newblock \doi{10.1016/j.cub.2020.08.052}.

\bibitem[Tagliazucchi et~al.(2016)Tagliazucchi, Chialvo, Siniatchkin, Amico,
  Brichant, Bonhomme, Noirhomme, Laufs, and Laureys]{Tagliazucchi2016}
Enzo Tagliazucchi, Dante~R. Chialvo, Michael Siniatchkin, Enrico Amico,
  Jean-Francois Brichant, Vincent Bonhomme, Quentin Noirhomme, Helmut Laufs,
  and Steven Laureys.
\newblock Large-scale signatures of unconsciousness are consistent with a
  departure from critical dynamics.
\newblock \emph{Journal of The Royal Society Interface}, 13\penalty0
  (114):\penalty0 20151027, 2016.
\newblock \doi{10.1098/rsif.2015.1027}.

\bibitem[Bellay et~al.(2015)Bellay, Klaus, Seshadri, and Plenz]{Bellay2015}
Timothy Bellay, Andreas Klaus, Saurav Seshadri, and Dietmar Plenz.
\newblock Irregular spiking of pyramidal neurons organizes as scale-invariant
  neuronal avalanches in the awake state.
\newblock \emph{eLife}, 4:\penalty0 e07224, jul 2015.
\newblock ISSN 2050-084X.
\newblock \doi{10.7554/eLife.07224}.

\bibitem[Hahn et~al.(2017)Hahn, Ponce-Alvarez, Monier, Benvenuti, Kumar,
  Chavane, Deco, and Frégnac]{Hahn2017}
Gerald Hahn, Adrian Ponce-Alvarez, Cyril Monier, Giacomo Benvenuti, Arvind
  Kumar, Frédéric Chavane, Gustavo Deco, and Yves Frégnac.
\newblock Spontaneous cortical activity is transiently poised close to
  criticality.
\newblock \emph{PLOS Computational Biology}, 13\penalty0 (5):\penalty0 1--29,
  05 2017.
\newblock \doi{10.1371/journal.pcbi.1005543}.

\bibitem[Priesemann et~al.(2013)Priesemann, Valderrama, Wibral, and
  Le~Van~Quyen]{Priesemann2013}
Viola Priesemann, Mario Valderrama, Michael Wibral, and Michel Le~Van~Quyen.
\newblock Neuronal avalanches differ from wakefulness to deep sleep –
  evidence from intracranial depth recordings in humans.
\newblock \emph{PLOS Computational Biology}, 9\penalty0 (3):\penalty0 1--14, 03
  2013.
\newblock \doi{10.1371/journal.pcbi.1002985}.

\bibitem[Fagerholm et~al.(2015)Fagerholm, Lorenz, Scott, Dinov, Hellyer,
  Mirzaei, Leeson, Carmichael, Sharp, Shew, and Leech]{Fagerholm2015}
Erik~D. Fagerholm, Romy Lorenz, Gregory Scott, Martin Dinov, Peter~J. Hellyer,
  Nazanin Mirzaei, Clare Leeson, David~W. Carmichael, David~J. Sharp,
  Woodrow~L. Shew, and Robert Leech.
\newblock Cascades and cognitive state: focused attention incurs subcritical
  dynamics.
\newblock \emph{Journal of Neuroscience}, 35\penalty0 (11):\penalty0
  4626--4634, 2015.
\newblock ISSN 0270-6474.
\newblock \doi{10.1523/JNEUROSCI.3694-14.2015}.

\bibitem[Parpura and Haydon(2000)]{Parpura8629}
Vladimir Parpura and Philip~G. Haydon.
\newblock Physiological astrocytic calcium levels stimulate glutamate release
  to modulate adjacent neurons.
\newblock \emph{Proceedings of the National Academy of Sciences}, 97\penalty0
  (15):\penalty0 8629--8634, 2000.
\newblock ISSN 0027-8424.
\newblock \doi{10.1073/pnas.97.15.8629}.

\bibitem[Christian and Huguenard(2013)]{RN62}
Catherine~A. Christian and John~R. Huguenard.
\newblock Astrocytes potentiate gabaergic transmission in the thalamic
  reticular nucleus via endozepine signaling.
\newblock \emph{Proceedings of the National Academy of Sciences}, 110\penalty0
  (50):\penalty0 20278--20283, 2013.
\newblock \doi{10.1073/pnas.1318031110}.

\bibitem[Mederos et~al.(2018)Mederos, González-Arias, and Perea]{RN63}
Sara Mederos, Candela González-Arias, and Gertrudis Perea.
\newblock Astrocyte–neuron networks: a multilane highway of signaling for
  homeostatic brain function.
\newblock \emph{Frontiers in Synaptic Neuroscience}, 10\penalty0 (45), 2018.
\newblock ISSN 1663-3563.
\newblock \doi{10.3389/fnsyn.2018.00045}.

\bibitem[Shigetomi et~al.(2016)Shigetomi, Patel, and Khakh]{RN64}
Eiji Shigetomi, Sandip Patel, and Baljit~S. Khakh.
\newblock Probing the complexities of astrocyte calcium signaling.
\newblock \emph{Trends in Cell Biology}, 26\penalty0 (4):\penalty0 300--312,
  2016.
\newblock ISSN 0962-8924.

\bibitem[Perea et~al.(2014)Perea, Sur, and Araque]{RN48}
Gertrudis Perea, Mriganka Sur, and Alfonso Araque.
\newblock Neuron-glia networks: integral gear of brain function.
\newblock \emph{Frontiers in Cellular Neuroscience}, 8\penalty0 (378), 2014.
\newblock ISSN 1662-5102.
\newblock \doi{10.3389/fncel.2014.00378}.

\bibitem[Araque et~al.(2014)Araque, Carmignoto, Haydon, Oliet, Robitaille, and
  Volterra]{RN67}
Alfonso Araque, Giorgio Carmignoto, Philip~G Haydon, Stéphane~HR Oliet,
  Richard Robitaille, and Andrea Volterra.
\newblock Gliotransmitters travel in time and space.
\newblock \emph{Neuron}, 81\penalty0 (4):\penalty0 728--739, 2014.
\newblock ISSN 0896-6273.

\bibitem[Lecun et~al.(1998)Lecun, Bottou, Bengio, and Haffner]{mnist_bib}
Y.~Lecun, L.~Bottou, Y.~Bengio, and P.~Haffner.
\newblock Gradient-based learning applied to document recognition.
\newblock \emph{Proceedings of the IEEE}, 86\penalty0 (11):\penalty0
  2278--2324, 1998.
\newblock \doi{10.1109/5.726791}.

\bibitem[Orchard et~al.(2015)Orchard, Jayawant, Cohen, and Thakor]{nmnist_bib}
Garrick Orchard, Ajinkya Jayawant, Gregory~K. Cohen, and Nitish Thakor.
\newblock Converting static image datasets to spiking neuromorphic datasets
  using saccades.
\newblock \emph{Frontiers in Neuroscience}, 9:\penalty0 437, 2015.
\newblock ISSN 1662-453X.
\newblock \doi{10.3389/fnins.2015.00437}.

\bibitem[Xiao et~al.(2017)Xiao, Rasul, and Vollgraf]{Xiao2017}
Han Xiao, Kashif Rasul, and Roland Vollgraf.
\newblock Fashion-mnist: a novel image dataset for benchmarking machine
  learning algorithms, 2017.

\bibitem[Morrison et~al.(2008)Morrison, Diesmann, and Gerstner]{Morrison2008}
Abigail Morrison, Markus Diesmann, and Wulfram Gerstner.
\newblock Phenomenological models of synaptic plasticity based on spike timing.
\newblock \emph{Biological Cybernetics}, 98\penalty0 (6):\penalty0 459--478,
  2008.
\newblock ISSN 1432-0770.
\newblock \doi{10.1007/s00422-008-0233-1}.

\bibitem[Zhang and Li(2019)]{Zhang2019}
Wenrui Zhang and Peng Li.
\newblock Information-theoretic intrinsic plasticity for online unsupervised
  learning in spiking neural networks.
\newblock \emph{Frontiers in Neuroscience}, 13:\penalty0 31, 2019.
\newblock ISSN 1662-453X.
\newblock \doi{10.3389/fnins.2019.00031}.

\bibitem[Maass et~al.(2004)Maass, Legenstein, and Bertschinger]{Maass_2004}
Wolfgang Maass, Robert Legenstein, and Nils Bertschinger.
\newblock Methods for estimating the computational power and generalization
  capability of neural microcircuits.
\newblock In \emph{Proceedings of the 17th International Conference on Neural
  Information Processing Systems}, NIPS'04, page 865–872, Cambridge, MA, USA,
  2004. MIT Press.

\bibitem[Wijesinghe et~al.(2019)Wijesinghe, Srinivasan, Panda, and
  Roy]{multi_liquid_lsm_2019}
Parami Wijesinghe, Gopalakrishnan Srinivasan, Priyadarshini Panda, and Kaushik
  Roy.
\newblock Analysis of liquid ensembles for enhancing the performance and
  accuracy of liquid state machines.
\newblock \emph{Frontiers in Neuroscience}, 13:\penalty0 504, 2019.
\newblock ISSN 1662-453X.
\newblock \doi{10.3389/fnins.2019.00504}.

\bibitem[Boedecker et~al.(2012)Boedecker, Obst, Lizier, Mayer, and
  Asada]{boedecker_2012}
Joschka Boedecker, Oliver Obst, Joseph~T Lizier, N~Michael Mayer, and Minoru
  Asada.
\newblock Information processing in echo state networks at the edge of chaos.
\newblock \emph{Theory Biosci.}, 131\penalty0 (3):\penalty0 205--13, 2012.
\newblock \doi{10.1007/s12064-011-0146-8}.

\bibitem[Ribeiro et~al.(2008)Ribeiro, Kauffman, Lloyd-Price, Samuelsson, and
  Socolar]{Ribeiro_2007}
Andre~S. Ribeiro, Stuart~A. Kauffman, Jason Lloyd-Price, Bj\"orn Samuelsson,
  and Joshua E.~S. Socolar.
\newblock Mutual information in random boolean models of regulatory networks.
\newblock \emph{Phys. Rev. E}, 77:\penalty0 011901, Jan 2008.
\newblock \doi{10.1103/PhysRevE.77.011901}.

\bibitem[Ku\ifmmode~\acute{s}\else \'{s}\fi{}mierz
  et~al.(2020)Ku\ifmmode~\acute{s}\else \'{s}\fi{}mierz, Ogawa, and
  Toyoizumi]{kus_2020}
\L{}ukasz Ku\ifmmode~\acute{s}\else \'{s}\fi{}mierz, Shun Ogawa, and Taro
  Toyoizumi.
\newblock Edge of chaos and avalanches in neural networks with heavy-tailed
  synaptic weight distribution.
\newblock \emph{Phys. Rev. Lett.}, 125:\penalty0 028101, Jul 2020.
\newblock \doi{10.1103/PhysRevLett.125.028101}.

\bibitem[van Vreeswijk and Sompolinsky(1996)]{Vreeswijk_1996}
C.~van Vreeswijk and H.~Sompolinsky.
\newblock Chaos in neuronal networks with balanced excitatory and inhibitory
  activity.
\newblock \emph{Science}, 274\penalty0 (5293):\penalty0 1724--1726, 1996.
\newblock \doi{10.1126/science.274.5293.1724}.

\bibitem[Ostojic(2014)]{Ostojic_2014}
Srdjan Ostojic.
\newblock Two types of asynchronous activity in networks of excitatory and
  inhibitory spiking neurons.
\newblock \emph{Nature Neuroscience}, 17:\penalty0 594--600, 2014.
\newblock ISSN 1546-1726.
\newblock \doi{10.1038/nn.3658}.

\bibitem[Rajan et~al.(2010)Rajan, Abbott, and Sompolinsky]{Rajan_2010}
Kanaka Rajan, L.~F. Abbott, and Haim Sompolinsky.
\newblock Stimulus-dependent suppression of chaos in recurrent neural networks.
\newblock \emph{Phys. Rev. E}, 82:\penalty0 011903, Jul 2010.
\newblock \doi{10.1103/PhysRevE.82.011903}.

\bibitem[Zhou et~al.(2019)Zhou, Zuo, and Jiang]{Zhou2019}
Bin Zhou, Yun-Xia Zuo, and Ruo-Tian Jiang.
\newblock Astrocyte morphology: diversity, plasticity, and role in neurological
  diseases.
\newblock \emph{CNS Neuroscience \& Therapeutics}, 25\penalty0 (6):\penalty0
  665--673, 2019.
\newblock \doi{10.1111/cns.13123}.

\bibitem[Zhao et~al.(2020)Zhao, Zeng, Zhang, Shi, and fei
  Zhao]{Zhao2020GLSNNAM}
Dongcheng Zhao, Yi~Zeng, Tielin Zhang, Mengting Shi, and Fei fei Zhao.
\newblock Glsnn: A multi-layer spiking neural network based on global feedback
  alignment and local stdp plasticity.
\newblock \emph{Frontiers in Computational Neuroscience}, 14, 2020.

\bibitem[Samadi et~al.(2017)Samadi, Lillicrap, and Tweed]{Samadi2017DeepLW}
Arash Samadi, T.~Lillicrap, and D.~Tweed.
\newblock Deep learning with dynamic spiking neurons and fixed feedback
  weights.
\newblock \emph{Neural Computation}, 29:\penalty0 578--602, 2017.

\bibitem[Zhang et~al.(2018{\natexlab{a}})Zhang, Zeng, Zhao, and
  Xu]{Zhang_2018_balsnn}
Tielin Zhang, Yi~Zeng, Dongcheng Zhao, and Bo~Xu.
\newblock Brain-inspired balanced tuning for spiking neural networks.
\newblock In \emph{Proceedings of the Twenty-Seventh International Joint
  Conference on Artificial Intelligence, {IJCAI-18}}, pages 1653--1659.
  International Joint Conferences on Artificial Intelligence Organization, 7
  2018{\natexlab{a}}.
\newblock \doi{10.24963/ijcai.2018/229}.

\bibitem[Diehl and Cook(2015)]{Diehl_2015}
Peter Diehl and Matthew Cook.
\newblock Unsupervised learning of digit recognition using
  spike-timing-dependent plasticity.
\newblock \emph{Frontiers in Computational Neuroscience}, 9:\penalty0 99, 2015.
\newblock ISSN 1662-5188.
\newblock \doi{10.3389/fncom.2015.00099}.

\bibitem[Mostafa(2018)]{Mostafa2018SupervisedLB}
Hesham Mostafa.
\newblock Supervised learning based on temporal coding in spiking neural
  networks.
\newblock \emph{IEEE Transactions on Neural Networks and Learning Systems},
  29:\penalty0 3227--3235, 2018.

\bibitem[Mirsadeghi et~al.(2021)Mirsadeghi, Shalchian, Kheradpisheh, and
  Masquelier]{Mirsadeghi_2021}
Maryam Mirsadeghi, Majid Shalchian, Saeed~Reza Kheradpisheh, and Timothée
  Masquelier.
\newblock Stidi-bp: Spike time displacement based error backpropagation in
  multilayer spiking neural networks.
\newblock \emph{Neurocomputing}, 427:\penalty0 131--140, 2021.
\newblock ISSN 0925-2312.
\newblock \doi{https://doi.org/10.1016/j.neucom.2020.11.052}.

\bibitem[O'Connor and Welling(2016)]{OConnor2016DeepSN}
Peter O'Connor and M.~Welling.
\newblock Deep spiking networks.
\newblock \emph{ArXiv}, abs/1602.08323, 2016.

\bibitem[Lee et~al.(2016)Lee, Delbruck, and Pfeiffer]{Lee_2016_backpropsnn}
Jun~Haeng Lee, Tobi Delbruck, and Michael Pfeiffer.
\newblock Training deep spiking neural networks using backpropagation.
\newblock \emph{Frontiers in Neuroscience}, 10:\penalty0 508, 2016.
\newblock ISSN 1662-453X.
\newblock \doi{10.3389/fnins.2016.00508}.

\bibitem[Wu et~al.(2018)Wu, Deng, Li, Zhu, and Shi]{Wu_2018_ssnbackprop}
Yujie Wu, Lei Deng, Guoqi Li, Jun Zhu, and Luping Shi.
\newblock Spatio-temporal backpropagation for training high-performance spiking
  neural networks.
\newblock \emph{Frontiers in Neuroscience}, 12:\penalty0 331, 2018.
\newblock ISSN 1662-453X.
\newblock \doi{10.3389/fnins.2018.00331}.

\bibitem[Kaiser et~al.(2020)Kaiser, Mostafa, and Neftci]{Kaiser_2020}
Jacques Kaiser, Hesham Mostafa, and Emre Neftci.
\newblock Synaptic plasticity dynamics for deep continuous local learning
  (decolle).
\newblock \emph{Frontiers in Neuroscience}, 14:\penalty0 424, 2020.
\newblock ISSN 1662-453X.
\newblock \doi{10.3389/fnins.2020.00424}.

\bibitem[Liu et~al.(2020{\natexlab{b}})Liu, Ruan, Xing, Tang, and
  Pan]{Liu2020EffectiveAO}
Qianhui Liu, Haibo Ruan, Dong Xing, H.~Tang, and Gang Pan.
\newblock Effective aer object classification using segmented
  probability-maximization learning in spiking neural networks.
\newblock In \emph{AAAI}, 2020{\natexlab{b}}.

\bibitem[Shrestha and Orchard(2018)]{Shrestha2018SLAYERSL}
S.~Shrestha and G.~Orchard.
\newblock Slayer: Spike layer error reassignment in time.
\newblock In \emph{NeurIPS}, 2018.

\bibitem[Zhang et~al.(2018{\natexlab{b}})Zhang, Zeng, Zhao, and
  Shi]{Zhang_2018}
Tielin Zhang, Yi~Zeng, Dongcheng Zhao, and Mengting Shi.
\newblock A plasticity-centric approach to train the non-differential spiking
  neural networks.
\newblock In \emph{Proceedings of the AAAI Conference on Artificial
  Intelligence}, 2018{\natexlab{b}}.

\bibitem[Hao et~al.(2020)Hao, Huang, Dong, and Xu]{HAO2020}
Yunzhe Hao, Xuhui Huang, Meng Dong, and Bo~Xu.
\newblock A biologically plausible supervised learning method for spiking
  neural networks using the symmetric stdp rule.
\newblock \emph{Neural Networks}, 121:\penalty0 387--395, 2020.
\newblock ISSN 0893-6080.
\newblock \doi{https://doi.org/10.1016/j.neunet.2019.09.007}.

\bibitem[Kheradpisheh et~al.(2020)Kheradpisheh, Mirsadeghi, and
  Masquelier]{Kheradpisheh_2007}
Saeed~Reza Kheradpisheh, Maryam Mirsadeghi, and Timoth{\'{e}}e Masquelier.
\newblock {BS4NN:} binarized spiking neural networks with temporal coding and
  learning.
\newblock \emph{CoRR}, abs/2007.04039, 2020.

\bibitem[Ju et~al.(2013)Ju, Xu, Chong, and VanDongen]{JU201339}
Han Ju, Jian-Xin Xu, Edmund Chong, and Antonius~M.J. VanDongen.
\newblock Effects of synaptic connectivity on liquid state machine performance.
\newblock \emph{Neural Networks}, 38:\penalty0 39--51, 2013.
\newblock ISSN 0893-6080.
\newblock \doi{https://doi.org/10.1016/j.neunet.2012.11.003}.

\bibitem[Ponulak and Kasi\'{n}ski(2010)]{Ponulak_2010}
Filip Ponulak and Andrzej Kasi\'{n}ski.
\newblock Supervised learning in spiking neural networks with resume: Sequence
  learning, classification, and spike shifting.
\newblock \emph{Neural Comput.}, 22\penalty0 (2):\penalty0 467–510, February
  2010.
\newblock ISSN 0899-7667.
\newblock \doi{10.1162/neco.2009.11-08-901}.

\bibitem[Urbanczik and Senn(2009)]{Urbanczik_2009}
Robert Urbanczik and Walter Senn.
\newblock {A Gradient Learning Rule for the Tempotron}.
\newblock \emph{Neural Computation}, 21\penalty0 (2):\penalty0 340--352, 02
  2009.
\newblock ISSN 0899-7667.
\newblock \doi{10.1162/neco.2008.09-07-605}.

\bibitem[Gütig and Sompolinsky(2006)]{Gutig_2006}
Robert Gütig and Haim Sompolinsky.
\newblock The tempotron: a neuron that learns spike timing–based decisions.
\newblock \emph{Nature Neuroscience}, 9\penalty0 (3):\penalty0 420--428, 2006.
\newblock \doi{10.1038/nn1643}.

\bibitem[Tang et~al.(2019{\natexlab{b}})Tang, Polykretis, Ivanov, Shah, and
  Michmizos]{Tang_2019_2}
Guangzhi Tang, Ioannis~E. Polykretis, Vladimir~A. Ivanov, Arpit Shah, and
  Konstantinos~P. Michmizos.
\newblock Introducing astrocytes on a neuromorphic processor: Synchronization,
  local plasticity and edge of chaos.
\newblock \emph{ACM Proceedings of 2019 Neuroinspired Computing Elements (NICE
  2019)}, 1\penalty0 (1):\penalty0 1--10, 2019{\natexlab{b}}.
\newblock \doi{arXiv:1907.01620}.

\bibitem[Roxin et~al.(2011)Roxin, Brunel, Hansel, Mongillo, and van
  Vreeswijk]{Roxin2011}
Alex Roxin, Nicolas Brunel, David Hansel, Gianluigi Mongillo, and Carl van
  Vreeswijk.
\newblock On the distribution of firing rates in networks of cortical neurons.
\newblock \emph{Journal of Neuroscience}, 31\penalty0 (45):\penalty0
  16217--16226, 2011.
\newblock ISSN 0270-6474.
\newblock \doi{10.1523/JNEUROSCI.1677-11.2011}.

\bibitem[Shew et~al.(2009)Shew, Yang, Petermann, Roy, and Plenz]{Shew2009}
Woodrow~L. Shew, Hongdian Yang, Thomas Petermann, Rajarshi Roy, and Dietmar
  Plenz.
\newblock Neuronal avalanches imply maximum dynamic range in cortical networks
  at criticality.
\newblock \emph{Journal of Neuroscience}, 29\penalty0 (49):\penalty0
  15595--15600, 2009.
\newblock ISSN 0270-6474.
\newblock \doi{10.1523/JNEUROSCI.3864-09.2009}.

\bibitem[Zenke et~al.(2013)Zenke, Hennequin, and Gerstner]{Zenke2013}
Friedemann Zenke, Guillaume Hennequin, and Wulfram Gerstner.
\newblock Synaptic plasticity in neural networks needs homeostasis with a fast
  rate detector.
\newblock \emph{PLOS Computational Biology}, 9\penalty0 (11):\penalty0 1--14,
  11 2013.
\newblock \doi{10.1371/journal.pcbi.1003330}.

\bibitem[Halassa and Haydon(2010)]{RN55}
Michael~M. Halassa and Philip~G. Haydon.
\newblock Integrated brain circuits: astrocytic networks modulate neuronal
  activity and behavior.
\newblock \emph{Annual Review of Physiology}, 72\penalty0 (1):\penalty0
  335--355, 2010.
\newblock \doi{10.1146/annurev-physiol-021909-135843}.

\bibitem[Benarroch(2005)]{RN54}
Eduardo~E. Benarroch.
\newblock Neuron-astrocyte interactions: partnership for normal function and
  disease in the central nervous system.
\newblock \emph{Mayo Clinic Proceedings}, 80\penalty0 (10):\penalty0
  1326--1338, 2005.
\newblock ISSN 0025-6196.

\bibitem[Theodosis et~al.(2008)Theodosis, Poulain, and Oliet]{RN53}
Dionysia~T. Theodosis, Dominique~A. Poulain, and Stéphane H.~R. Oliet.
\newblock Activity-dependent structural and functional plasticity of
  astrocyte-neuron interactions.
\newblock \emph{Physiological Reviews}, 88\penalty0 (3):\penalty0 983--1008,
  2008.
\newblock \doi{10.1152/physrev.00036.2007}.

\end{thebibliography}

\newpage
\appendix
\section{Appendix}
\subsection{Datasets} \label{sup_datasets}
We tested models on MNIST \citeappend{mnist_bib_0}, its temporal, event-driven version, N-MNIST \citeappend{nmnist_bib_0}, and Fashion-MNIST \citeappend{fmnist_bib_0}. We modified the original $60,000$/$10,000$ train/test split to $50,000$/$10,000$/$10,000$ train/validate/test split, by partitioning away the last $10,000$ training samples to the validation set. We normalized and rescaled each $28\times 28$ MNIST and Fashion-MNIST image to $0-1$ range, which we Poisson encoded into the spiking activity of input neurons. For N-MNIST, we treated all discrete events the same way and transformed each image into $300\times 68\times 34$ matrix, with the first dimension being temporal. Using first $250$ timesteps, we converted each event at each timestep into a spike in the corresponding input neuron.

\subsection{Neuron Parameters} \label{sup_neuron_params}
The LIF neuron parameters we used in all networks are shown in Table \ref{table_lif_params}.

\begin{table}[b]
\renewcommand{\thetable}{S1}
  \caption{LIF neuron parameters}
  \centering
  \begin{tabular}{lll}
    \toprule
    \cmidrule(r){1-2}
    Parameter name     & Description     & Value \\
    \midrule
    $\theta$      & membrane potential threshold       & $20.0$     \\
    $\tau_{v}$    & membrane potential time constant   & $64.0$      \\
    $\tau_{u}$    & synaptic conductance time constant & $1.0$  \\
    $b$           & membrane potential bias           & $0.0$  \\
    \bottomrule
  \end{tabular}
  \label{table_lif_params}
\end{table}

\subsection{Liquid Connectivity Parameters} \label{sup_connectivity}
The parameters we used in the distance based connection probability function, (\ref{eq:connection_probability}), depended on the connection type. Connection types were determined by the pre- and post-synaptic neurons, which resulted in 4 types of connections:
\begin{enumerate}
    \item $EE$: excitatory to excitatory
    \item $EI$: excitatory to inhibitory
    \item $II$: inhibitory to inhibitory
    \item $IE$: inhibitory to excitatory
\end{enumerate}
For each connection type $[EE,EI,II,IE]$, parameter $C$ values were $[0.2,0.1,0.3,0.05]$. For all connection types, $\lambda=3.0$.

\subsection{Neuron-astrocyte connection weight}\label{sup_astro}
\begin{figure}
\renewcommand{\thefigure}{S1}
  \centering
    \includegraphics[width=1.0\linewidth]{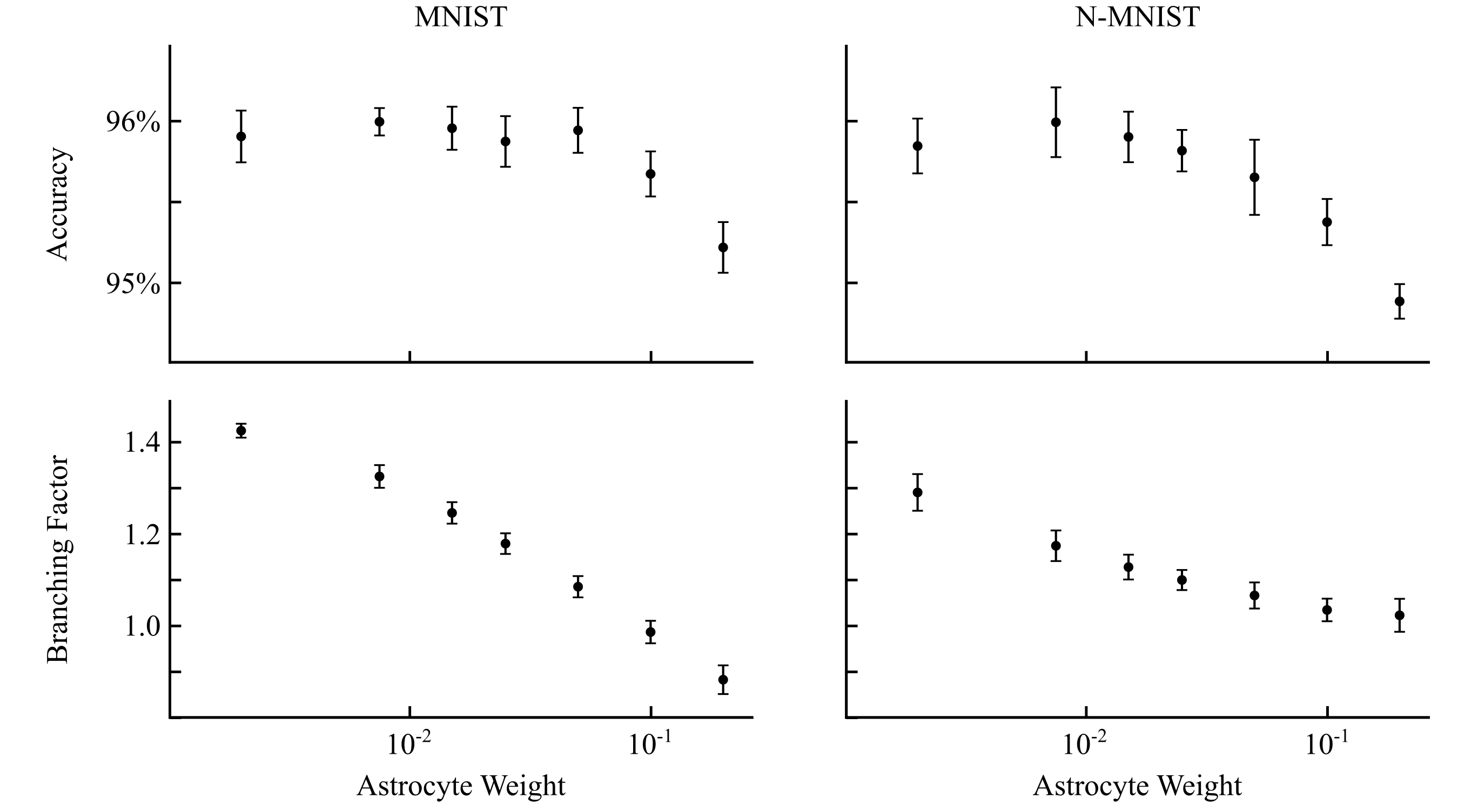}
  \caption{{\bf{Neuron-astrocyte connection weight impacts liquid dynamics and NALSM accuracy.}} ( {\bf{Top}} ) NALSM accuracy shown with respect to neuron-astrocyte connection weight for MNIST and N-MNIST. ( {\bf{Bottom}} ) NALSM liquid dynamics shown as a function of neuron-astrocyte connection weight for MNIST and N-MNIST. Data points are averaged over 10 random networks. Error bars are standard deviation.}
  \label{supplementay_figure_astro_tuning}
\end{figure}

The weight of neuron-astrocyte connections, $w_{astro}$ in (\ref{eq:lim_astro}), impacted both liquid dynamics and NALSM accuracy. Controlling the responsiveness of the LIM astrocyte to liquid neuron activity, larger $w_{astro}$ resulted in lower branching factor, and vice versa. For both datasets, accuracy peaked in the vicinity of $w_{astro}=0.01$ with slightly super-critical branching factor of $\approx 1.3$ for MNIST and $\approx 1.2$ for N-MNIST (Fig. \ref{supplementay_figure_astro_tuning}).

\subsection{LSM+AP-STDP model}\label{sup_apstdp}
We implemented AP-STDP from \citeappend{apstdp_bib_0} on top of LSM+STDP model by making STDP weight changes conditionally dependent on neuronal activity. Specifically, we implemented rule (4) from \citeappend{apstdp_bib_0}, with $p=1.0$. The spiking rate of each neuron $i$, $C_{i}$, was approximated using (3) from \citeappend{apstdp_bib_0}, with $\tau_{C}=1000$ $ms$.
Parameters $C_{\theta}$ and $\Delta C$ set the neuronal activity range in which STDP changes were enforced. We hand-tuned parameters $C_{\theta}$ and $\Delta C$ for each specific network and dataset to maximize the validation accuracy of LSM+AP-STDP model. 
We used the same initialization process as was used for LSM+STDP model, with two exceptions 1) weights were set to $1.0$ prior to initialization, and 2) STDP synaptic weight changes were conditioned on neuron activity ranges using $C_{\theta}$ and $\Delta C$. As with LSM+STDP model, the liquid's weights were fixed during spike generation phase.

\subsection{Evidence for edge-of-chaos dynamics in NALSM}\label{appendix_edge_of_chaos}
\begin{figure}
\renewcommand{\thefigure}{S2}
  \centering
    \includegraphics{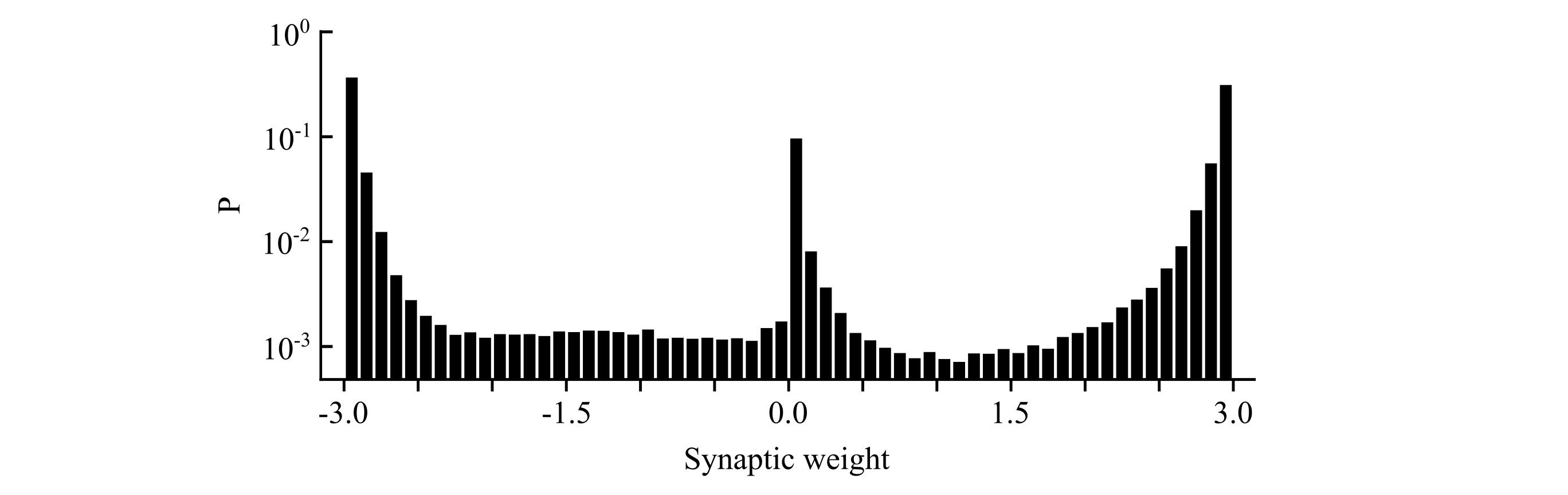}
  \caption{{\bf{Initialized NALSM synaptic weights.}} Distribution of IL and LL synaptic weights after NALSM liquid initialization.}
  \label{sup_fig_input_liquid_weights}
\end{figure}

Here, we provide evidence suggesting that NALSM's  slightly super-critical branching dynamics (Fig. \ref{results_fig_1}) corresponded to the edge-of-chaos. First, NALSM exhibited coexistence of small and large synaptic weights, which is necessary for chaotic activity in spiking networks \citeappend{kus_2020_0}. NALSM had concentrations of near-maximum excitatory weights and near-zero weights, with weights also covering the full range in between these extremes. Inhibitory weights exhibit the same kind of bimodal distribution (Fig. \ref{sup_fig_input_liquid_weights}). 

Second, NALSM exhibited excitation/inhibition (E/I) balance, which is thought to be necessary for existence of deterministic chaos \citeappend{Vreeswijk_1996_0,Krauss_2019_0}. We used three different methods to evaluate E/I balance. First, we confirmed synaptic weight E/I balance, $W_{E/I}$, in initialized NALSM liquids, which was found to align with edge-of-chaos dynamics in \citeappend{Krauss_2019_0} and was evaluated as: 
\begin{equation}\label{eq:300}
W_{E/I} = \frac{n_{w>0}-n_{w<0}}{n_{w\neq0}}\\
\end{equation}
where $n_{w>0}$, $n_{w<0}$, and $n_{w\neq0}$ are total number of IL and LL synaptic weights that are positive, negative, and non-zero, respectively. Indicative of E/I balance, we obtained $W_{E/I} = -0.0029\pm0.018$ averaged over all NALSM initializations on both MNIST and N-MNIST ($W_{E/I}$ ranges from $-1$ to $1$, with $0$ representing perfect E/I balance). Second, we measured the difference in spiking rates between liquid excitatory and liquid inhibitory neuron populations by evaluating:
\begin{equation}\label{eq:301}
f_{E/I} = \frac{|\hat{f}_{e}-\hat{f}_{i}|}{\hat{f}_{l}}\\
\end{equation}
where $\hat{f}_{e}$, $\hat{f}_{i}$, and $\hat{f}_{l}$ are the average spiking rate of excitatory liquid neurons, inhibitory liquid neurons, and all liquid neurons, respectively. Averaged over all NALSM network initializations and both MNIST and N-MNIST datasets, we obtained $f_{E/I} = 0.074\pm0.083$, which was indicative of E/I balance since $f_{E/I}$ ranges from $0$ to $1$, with $0$ representing perfect E/I balance. Finally, we measured the net current received by each neuron at each timestep from all active input and liquid neurons. Averaged over $100$ different MNIST input samples, net current received by each neuron was $- 0.99\pm 4.91$. The near $0$ average net current combined with its large standard deviation suggests that excitatory and inhibitory inputs were balanced and that neurons were primarily driven by network fluctuations. This is believed to give rise to the irregular activity observed in the brain \citeappend{Ostojic_2014_0} and has been associated with deterministic chaos \citeappend{Vreeswijk_1996_0} (shown in Fig. 2 A in \citeappend{Vreeswijk_1996_0}).

\begin{figure}
\renewcommand{\thefigure}{S3}
  \centering
    \includegraphics{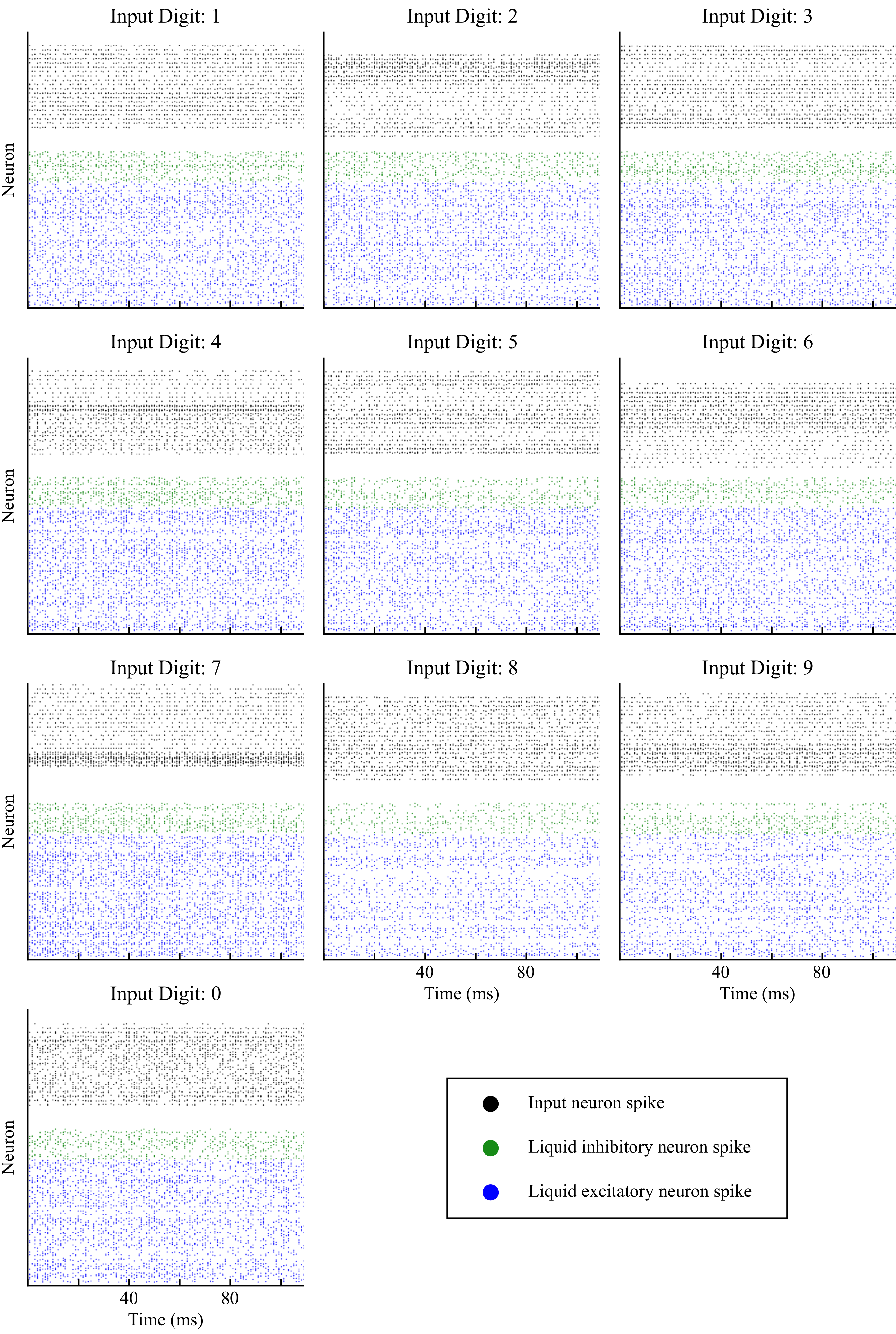}
  \caption{{\bf{NALSM network spike activity.}} For each input sample class, a raster plot shows spike activity of input (black), liquid inhibitory (green), and liquid excitatory (blue) neurons for a $100$ $ms$ duration.}
  \label{sup_fig_input_liquid_spike_raster}
\end{figure}

\begin{figure}
\renewcommand{\thefigure}{S4}
  \centering
    \includegraphics{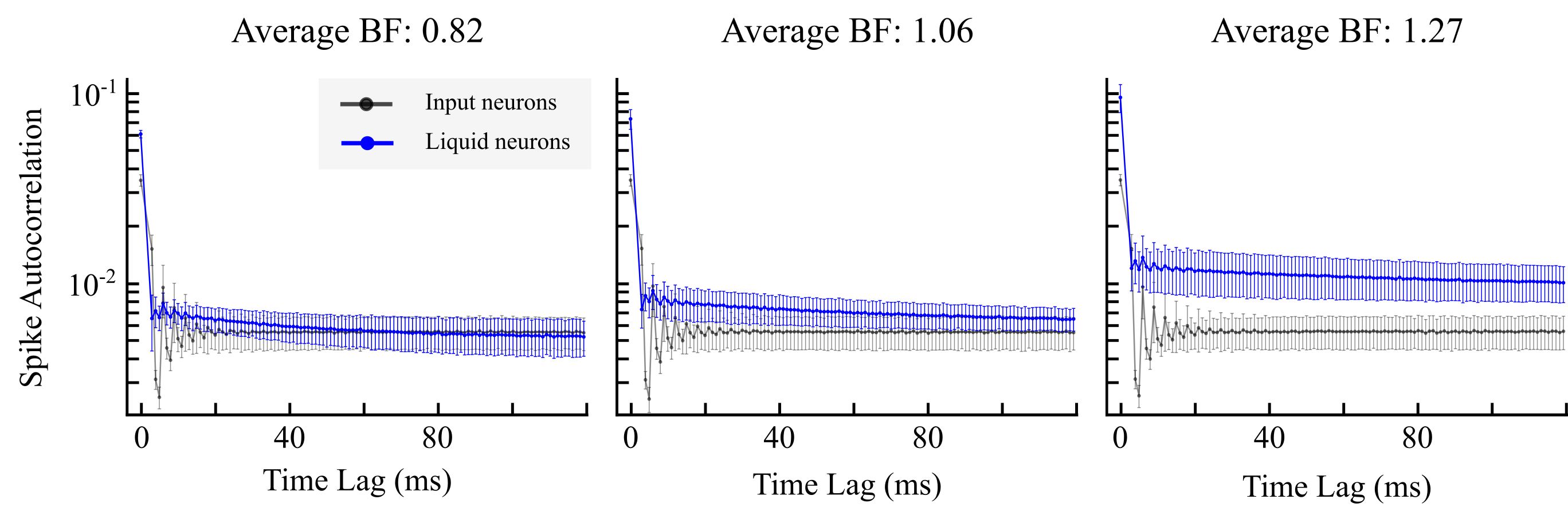}
  \caption{{\bf{Spike autocorrelation versus branching factor dynamics.}} Spike autocorrelation as a function of lag time for sub-critical (left), near-critical (middle), and super-critical (right) branching factor dynamics. Spike autocorrelation was computed using equation (\ref{eq:302}) on input (gray) and liquid (blue) neuron spike trains. Data points are averaged over 100 MNIST input samples. Error bars are standard deviation.}
  \label{sup_fig_input_liquid_autocorrelation}
\end{figure}

Indeed, NALSM also exhibited spiking activity that was irregular across the network and across time (Fig. \ref{sup_fig_input_liquid_spike_raster}). Evidence for chaotic activity was further confirmed by autocorrelation analysis performed on neuronal spike trains generated during generation of spike counts for output layer training (See \ref{methods__output layer}). Specifically, spike autocorrelation function, $A_{spikes}(\tau_{auto})$, was computed as:
\begin{equation}\label{eq:302}
A_{spikes}(\tau_{auto}) = \frac{1}{NT}\sum_{i=1}^{N}\sum_{t=1}^{T} \sigma_{i}(t+\tau_{auto})\sigma_{i}(t)
\end{equation}
where $N=1000$ liquid neurons, $T=125$ $ms$ is the duration of neuronal spike trains, $\sigma_{i}$ is the spike train of neuron $i$. As the branching factor became increasingly greater than $1.0$, the decay of liquid neuron spike autocorrelation functions became broader and increased in magnitude (Fig. \ref{sup_fig_input_liquid_autocorrelation}). Alternatively, when the branching factor became progressively less than $1.0$, decay of liquid neuron spike autocorrelation functions was narrower and magnitudes were marginally greater than that of input neuron spike autocorrelation functions. As expected, spike train autocorrelation functions of input neurons remained flat showing no decay with respect to lag time. This suggested that the transition from a sub-critical to a super-critical branching factor possibly corresponded to a transition to chaos in NALSM's spiking rate dynamics \citeappend{Ostojic_2014_0,Rajan_2010_0}.

\subsection{NALSM performance with respect to liquid size} \label{appendix_lsm_sizes}
\begin{figure}
\renewcommand{\thefigure}{S5}
  \centering
    \includegraphics{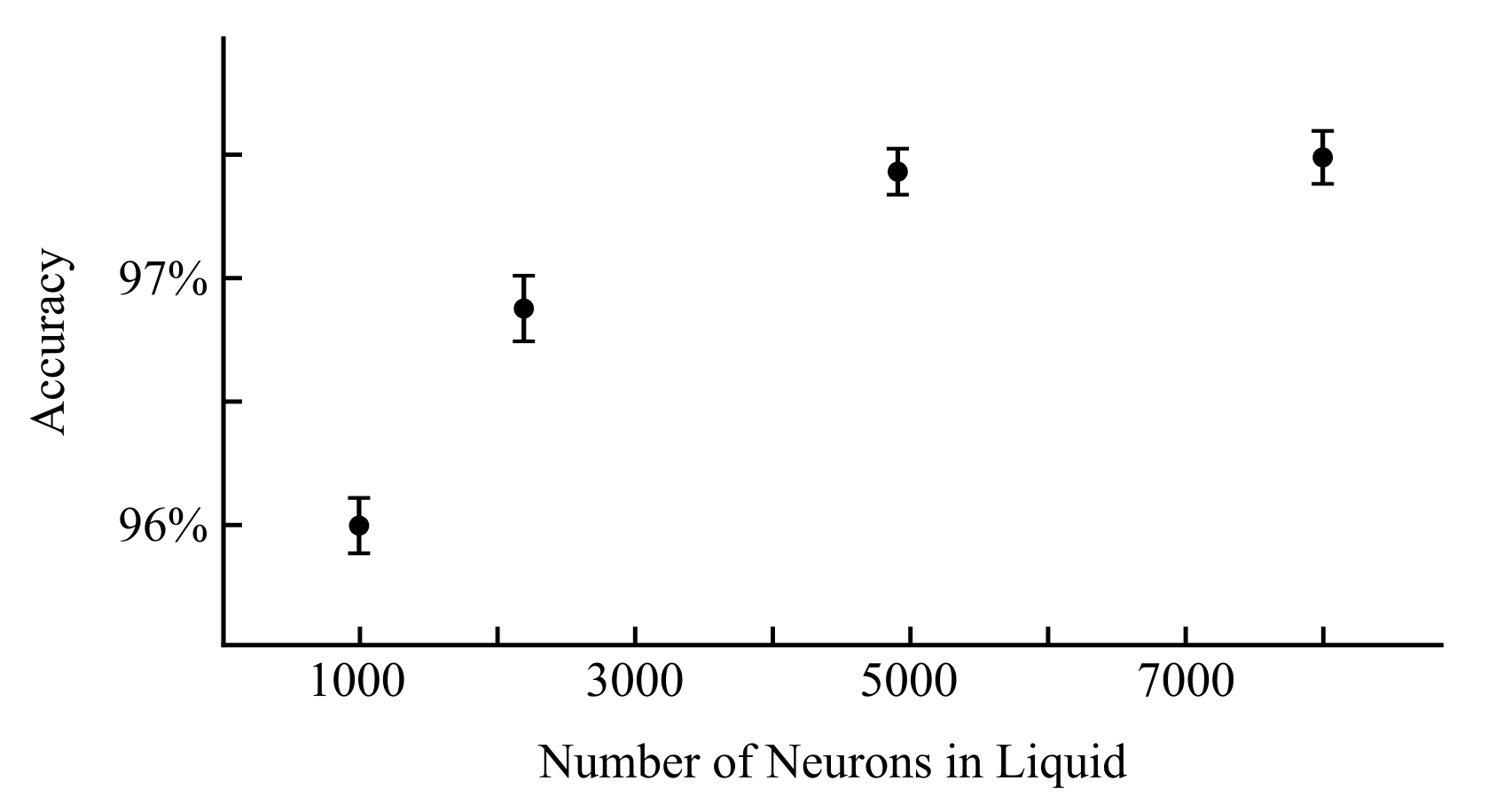}
  \caption{{\bf{NALSM accuracy increases with liquid size.}} NALSM accuracy shown with respect to number of neurons in the liquid. Data points are averaged over 5 random networks. Error bars are standard deviation.}
  \label{sup_fig_liquid_size}
\end{figure}

NALSM performance increased with the number of neurons in the liquid, saturating at approximately $8,000$ neurons (Fig. \ref{sup_fig_liquid_size})

\subsection{Number of plastic parameters in NALSM} \label{appendix_number_plastic_parameters}
For NALSM, we counted all IL, LL, and LO connections as either plastic with STDP or trainable with gradient descent. For NALSM$8000$, the number of LO connections was constant at $80,000$. The number of IL, LL connections varied based on the randomly generated liquid. The average number of total plastic/trainable connections for NALSM$8000$ trained on MNIST was $1,199,406.70 \pm 453.47$ with a maximum(minimum) of $1,200,105$($1,198,916$). For N-MNIST, the average was $3,033,045.40 \pm 268.70$ with a maximum(minimum) of $3,033,446$($3,032,737$). The significant difference in the number of plastic connections used for MNIST and N-MNIST training was due to the $\approx 3$ times larger input layer needed for N-MNIST.

\subsection{Added computational cost of the LIM astrocyte model} \label{appendix_computational_cost}
Our proposed method adds a negligible computational cost to the LSM. Specifically, we used a single astrocyte unit with the same functional form as the LIF neuron, making it $0.01\%$ of all the LIF neurons used in NALSM8000 (we used a total of $8,784$ input and liquid neurons for MNIST). In terms of connections, we used $8,784$ neuron-astrocyte connections, which was $0.78\%$ of the number of neuron-neuron connections (we used $1,119,407$ input-liquid and liquid-liquid connections). Further, we showed in Fig. \ref{results_fig_3_ABLATION} that even with $90\%$ of neuron-astrocyte connections removed, NALSM still maintains a performance advantage versus LSM+AP-STDP and LSM models; in which case only $878$ neuron-astrocyte connections are used or $0.078\%$ of neuron-neuron connections. Finally, fixed neuron-astrocyte connections are computationally less expensive than the plastic neuron-neuron connections, since the ms-precision STDP mechanism (Eqs. \ref{eq:STDP_W_CHANGE}, \ref{eq:stdp_pre_trace}, \ref{eq:stdp_post_trace}) adds extra computations on top of each neuronal connection that does not exist in the neuron-astrocyte connections.

\subsection{Curve Fitting} \label{appendix_curve_fits}
We fit $2$nd and $3$rd degree polynomial functions. All polynomial fit parameters and residual sum values are shown in Table \ref{table_poly_curve_fit_params}.

\begin{table}
\renewcommand{\thetable}{S2}
  \caption{Polynomial Curve Fitting Parameters}
  \centering
  \begin{tabular}{llll}
    \toprule
    Figure/Plot     & Degree  & Coefficients & Residuals Sum \\
    \midrule
    Fig. \ref{methods_fig_0} B & 3 & $(0.1143, -0.7563, 1.7492, -0.0658)$ & $0.00143$ \\
    Fig. \ref{results_fig_1} C MNIST   & 2 & $(-0.9151, 2.7587, -0.6077)$ & $0.00229$ \\
    Fig. \ref{results_fig_1} C N-MNIST & 2 & $(0.3000, 0.6104, 0.0752)$ & $0.00050$ \\
    Fig. \ref{results_fig_3_ABLATION} MNIST   & 3 & $(0.0187, -0.0344, 0.0209, 0.9561)$ & $0.000000422$ \\
    Fig. \ref{results_fig_3_ABLATION} N-MNIST & 3 & $(0.0044, -0.0084, 0.0060, 0.9585)$ & $0.000000477$ \\
    \bottomrule
  \end{tabular}
  \label{table_poly_curve_fit_params}
\end{table}

\subsection{Hardware} \label{compute_resources}
We used Tesla K$80$ GPU to train all models.

\newpage
\bibliographystyleappend{unsrtnat}
\bibliographyappend{bib_append}

\newpage

\end{document}